\documentclass[letterpaper, 10 pt, conference]{ieeeconf}  

\IEEEoverridecommandlockouts                              

\usepackage{graphics} 
\usepackage{epsfig} 
\usepackage{mathptmx} 
\usepackage{times} 
\usepackage{amsmath} 
\usepackage{amssymb}  
\usepackage{acro}
\usepackage{multirow}
\usepackage{tablefootnote} 
\usepackage{booktabs} 
\usepackage{bbding} 
\usepackage{subcaption} 
\usepackage{tabularx}
\usepackage{bbding}
\usepackage{xcolor} 
\usepackage{xcolor}

\DeclareAcronym{bbox}{
short=bbox,
long=bounding box
}

\DeclareAcronym{GT}{
short = GT,
long = ground truth
}

\DeclareAcronym{TPV}{
short=TPV,
long=tri-perspective view
}

\DeclareAcronym{LSS}{
short=LSS,
long=lift splat shot
}

\title{\LARGE \bf
Model Quantization and Hardware Acceleration for Vision Transformers: A Comprehensive Survey
}

\author{Dayou Du$^{1}$, Gu Gong$^{1}$, Xiaowen Chu$^{1\dagger}$%
\thanks{$^{1}$The Hong Kong University of Science and Technology (Guangzhou) \{ddu487, ggong504\}@connect.hkust-gz.edu.cn, xwchu@ust.hk}%
\thanks{$^{\dagger}$Corresponding author.}%
}

\begin{document}

\maketitle
\thispagestyle{empty}
\pagestyle{empty}


\begin{abstract}
Vision Transformers (ViTs) have recently garnered considerable attention, emerging as a promising alternative to convolutional neural networks (CNNs) in several vision-related applications. However, their large model sizes and high computational and memory demands hinder deployment, especially on resource-constrained devices. This underscores the necessity of algorithm-hardware co-design specific to ViTs, aiming to optimize their performance by tailoring both the algorithmic structure and the underlying hardware accelerator to each other's strengths. Model quantization, by converting high-precision numbers to lower-precision, reduces the computational demands and memory needs of ViTs, allowing the creation of hardware specifically optimized for these quantized algorithms, boosting efficiency. This article provides a comprehensive survey of ViTs quantization and its hardware acceleration. We first delve into the unique architectural attributes of ViTs and their runtime characteristics. Subsequently, we examine the fundamental principles of model quantization, followed by a comparative analysis of the state-of-the-art quantization techniques for ViTs. Additionally, we explore the hardware acceleration of quantized ViTs, highlighting the importance of hardware-friendly algorithm design. In conclusion, this article will discuss ongoing challenges and future research paths. We consistently maintain the related open-source materials at https://github.com/DD-DuDa/awesome-vit-quantization-acceleration.
\end{abstract}
\section{\textbf{Introduction}}

\begin{figure*}[htbp]
    \centering
    \includegraphics[width=0.9\linewidth]{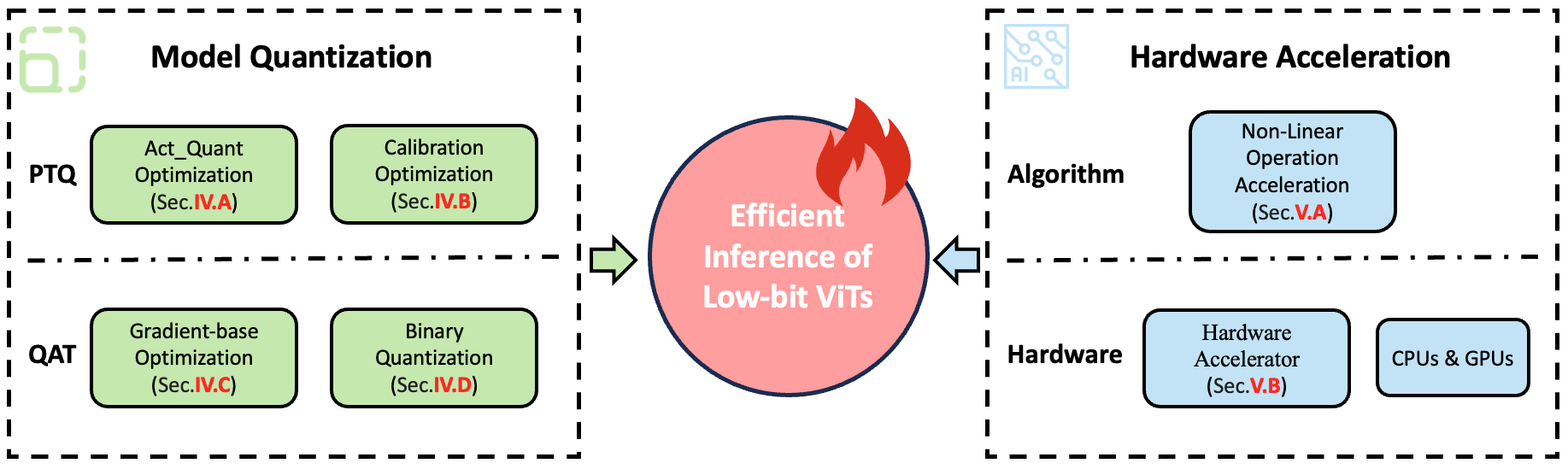}  
    \caption{Overview diagram for the survey on effective low-bit ViTs Inference.}
    \label{fig:overview}
\end{figure*}


In the realm of computer vision, Convolutional Neural Networks (CNNs) have historically been the cornerstone, demonstrating remarkable efficacy across a plethora of tasks \cite{cnn_survey}. However, the landscape began to shift with the advent of the Transformer architecture, which, after its resounding success in natural language processing (NLP) \cite{attention_is_all,bert,gpt3}, was adapted for computer vision in the form of Vision Transformers (ViTs) \cite{vit,deit,swin}. The pivotal feature of ViTs, self-attention, allows the model to contextually analyze visual data by learning intricate relationships between elements within a sequence of image tokens. This ability to grasp the broader context and the interdependencies within an image has propelled the rapid development of Transformer-based models in vision, subsequently establishing them as the new backbone for a diverse range of tasks, including image classification \cite{swin}, object detection \cite{detr}, image generation \cite{image_generation}, autonomous driving \cite{bevformer}, and visual question answering \cite{vision_answer}, showcasing their versatility and transformative impact in computer vision, as reviewed in \cite{vit_survey}.

Despite their remarkable capabilities, ViTs face challenges due to their inherently large model sizes and the quadratic increase in computational and memory demands posed by the self-attention mechanism, especially as image resolution grows. This combination significantly hinders deployment on devices with constrained computational and memory resources, particularly in real-time applications such as autonomous driving \cite{autonomous_survey} and virtual reality \cite{vr_vit}, where fulfilling low latency requirements and producing a high-quality user experience are crucial. This underscores the pressing need for advancements in model compression techniques such as pruning \cite{prune_vit}, quantization \cite{liu_ptq}, knowledge distillation \cite{kd_vit}, and low-rank factorization \cite{rank_vit}. Moreover, the rapid adoption of ViTs can be attributed not only to algorithmic innovations and data availability but also to enhancements in processor performance. While CPUs and GPUs offer broad computing versatility, their inherent flexibility can lead to inefficiencies. Given the repetitive yet distinct operations characteristic of ViTs, there is a clear opportunity for leveraging specialized hardware designed to optimize data reuse, thereby enhancing efficiency in ViT deployments.

Quantization, a technique that maps higher precision into lower precision, has been successful in facilitating lightweight and computationally efficient models, enhancing the interaction between algorithms and hardware \cite{gholami2022survey}. Various techniques have been specifically designed for ViTs on the algorithm side. These aim to maintain the accuracy of the application after compressing data to lower bitwidth. Some of these techniques are designed to be more hardware-friendly, taking into account existing architectures such as the GPU INT8/FP8 Tensorcore \cite{i-vit, hopperbenchmarking}. On the hardware side, optimization of high-level quantization algorithms drives the design of more efficient processors, potentially incorporating more efficient data reuse modules for parallel processing of lower-bit data \cite{huang2023integer, sole}. This co-design of algorithm and hardware is a common approach in the development of modern hardware accelerators, improving their performance significantly.

However, the vast array of related works published in recent years has made it challenging for beginners to obtain a comprehensive overview and clear comparative results. Moreover, some methods that simulate algorithm designs without considering the actual hardware may result in unexpectedly poor accuracy when deployed \cite{practical}. There is a pressing need for a comprehensive survey that summarizes, analyzes, and compares these methodologies. This article endeavors to fill this gap by providing an extensive review of ViTs quantization and its hardware acceleration. Specifically, we delve into the nuanced challenges of ViTs quantization from both algorithmic and hardware perspectives, offering a vertical comparison across different quantization approaches, as illustrated in Fig.\textcolor{red}{\ref{fig:overview}}. Additionally, we illustrate advanced hardware design solutions and speculate on future trends and potential opportunities. In contrast to recent surveys—some focusing on various efficient techniques without hardware consideration \cite{survey_efficient_vit,survey_model_compression}, some on inference optimization with limited algorithmic detail \cite{fullstack}, and others offering a broad overview of model compression mainly for large language models \cite{survey_llm, survey_llm2}—this article presents detailed descriptions and comparisons, addressing the interplay between algorithms and hardware in a synergistic manner, thus providing a clearer, more structured insight into the domain of ViTs quantization.

The organization of this paper is summarized as follows. Sec.\textcolor{red}{\ref{sec:vit}} delves into the architecture of Vision Transformers, presenting its variants and analyzing their runtime characteristics and bottlenecks with profiling. Sec.\textcolor{red}{\ref{sec:fund_quant}} elucidates the fundamental principles of model quantization. Subsequently, Sec.\textcolor{red}{\ref{sec:quant}} examines the pressing challenges associated with the quantization of ViTs and offers a comparative review of the performance of previous methodologies. Sec.\textcolor{red}{\ref{sec:hardware}} explores the range of methods available for hardware acceleration. Finally, Sec.\textcolor{red}{\ref{sec:conclusion}} provides a summary of the paper, highlighting potential opportunities and challenges.
\section{\textbf{Vision Transformers Model Architecture and Performance Analysis}}
\label{sec:vit}

The Vision Transformers  (ViTs) \cite{vit}, utilizing the self-attention mechanism to grasp ``long-range" relationships in image sequences, has recently achieved remarkable success across a variety of computer vision tasks, establishing itself as a versatile vision backbone \cite{vit_survey}. This section begins with an overarching overview of ViTs' architecture, delving into its various modules and operations in Sec.\textcolor{red}{\ref{subsec:overview_vit}}. Following this, we explore the evolution and variants of ViTs in Sec.\textcolor{red}{\ref{subsec:variant_vits}}. We conclude with an analysis of different operations' impact, utilizing the roofline model, as detailed in Sec.\textcolor{red}{\ref{subsec:roofline}}.

\subsection{\textbf{Overview of Vision Transformer Architecture}}
\label{subsec:overview_vit}

\begin{figure*}[htbp]
    \centering
    \includegraphics[width=0.9\linewidth]{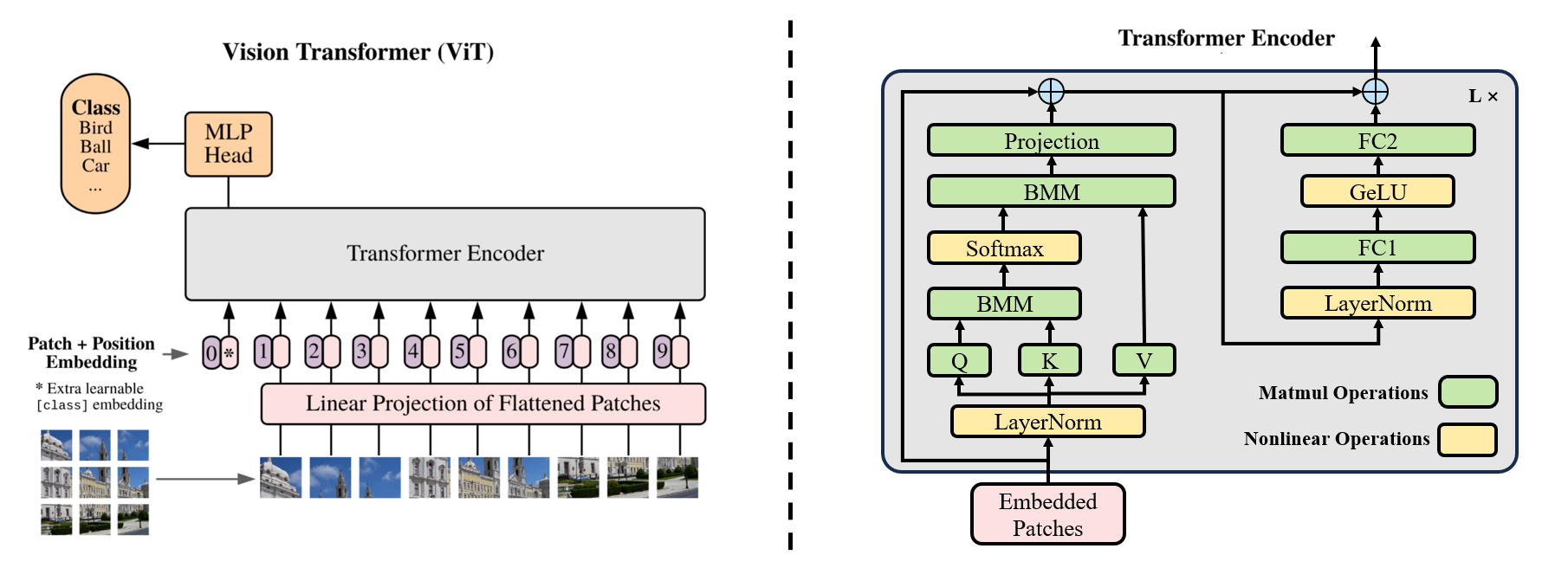}  
    \caption{Architecture of the Vision Transformer (ViT): The left illustrates the process of image division and positional embedding, while the right panel delineates the standard encoder architecture with its various operations, as detailed in \cite{vit}. The abbreviation `BMM' refers to batch matrix multiplication.}
    \label{fig:vit-arch}
\end{figure*}

The ViTs architecture is delineated in Fig.\textcolor{red}{\ref{fig:vit-arch}}, beginning by segmenting the input image into patches, which are then transformed into a linear sequence and supplemented with a class token. This sequence, inclusive of the class token, is then equipped with positional embeddings and processed through the Transformer Encoder layers for feature encoding. The process culminates with a fully-connected layer, termed the ``MLP Head", for classification purposes. The Transformer Encoder's functionality, illustrated in the right panel of Fig.\textcolor{red}{\ref{fig:vit-arch}}, encompasses both a Multi-Head Attention (MHA) and a Feed-Forward (FFN) module, and each of which is followed by a Layer Normalization (LayerNorm) operation and a residual connection.

The MHA module first projects the image sequence by multiplying it through distinct weight matrices (\(W^Q\), \(W^K\), and \(W^V\)), generating query (\(Q\)), key (\(K\)), and value (\(V\)) activation. The self-attention mechanism is then applied as follows:
\begin{equation}
\mathrm{Attention}(Q,K,V) = \mathrm{softmax}\left(\frac{QK^T}{\sqrt{d_k}}\right)V,
\end{equation}
where $d_k$ represents the dimensionality of the key vectors. The MHA aggregates information from multiple representation subspaces, synthesizing the outputs from different heads into a unified representation:
\begin{equation}
\text{MultiHead}(Q, K, V) = \text{Concat}(\text{head}_1, \ldots, \text{head}_h)W^O,
\label{eq:head}
\end{equation}
with each head defined as:
\begin{equation}
\text{head}_i = \text{Attention}(QW_i^Q, KW_i^K, VW_i^V).
\end{equation}

The FFN module, which includes two dense layers activated by GELU \cite{gelu}, processes each token individually, augmenting the model's capacity to understand intricate functions:
\begin{equation}
\mathrm{FFN}(x)=GELU(xW_1+b_1)W_2+b_2.
\end{equation}

In summary, the MHA encompasses six linear operations, including four weight-to-activation transformations (\(W^Q\), \(W^K\), \(W^V\), and \(W^O\) projections) and two activation-to-activation transformations (\(Q \times K^T\) and \(\text{Out}_{\text{softmax}} \times V\)). In contrast, the FFN comprises two linear projections (\(W_1\) and \(W_2\)). Non-linear operations like Softmax, LayerNorm, and GELU, though less prevalent, present computational challenges on conventional hardware due to their complexity, potentially restricting the enhancement of end-to-end transformer inference \cite{softermax}.

\subsection{\textbf{Variant ViTs}}
\label{subsec:variant_vits}

ViTs has been a foundational architecture for subsequent Transformer-based models in image processing tasks, known for its robustness in handling various scales and complexities in images. Building upon the strengths of ViT, DeiT \cite{deit} was introduced, which retains the original architecture but is specifically designed to be more data-efficient. By incorporating a novel teacher-student strategy tailored for Transformers, which includes the use of distillation tokens, DeiT can effectively train on smaller datasets without a substantial loss in performance, demonstrating the model's adaptability to data constraints.

Furthermore, Swin-Transformer \cite{swin} saw another significant advancement. This model enhances the ViT framework by integrating a hierarchical structure with shifted windows, a design choice that substantially improves the model's ability to capture local features while still maintaining a global context. 

For more information about this topic, see \cite{vit_survey,vit_survey2} for a more comprehensive survey of ViTs.

\begin{figure}[h]
    \centering
    \includegraphics[width=0.85\linewidth]{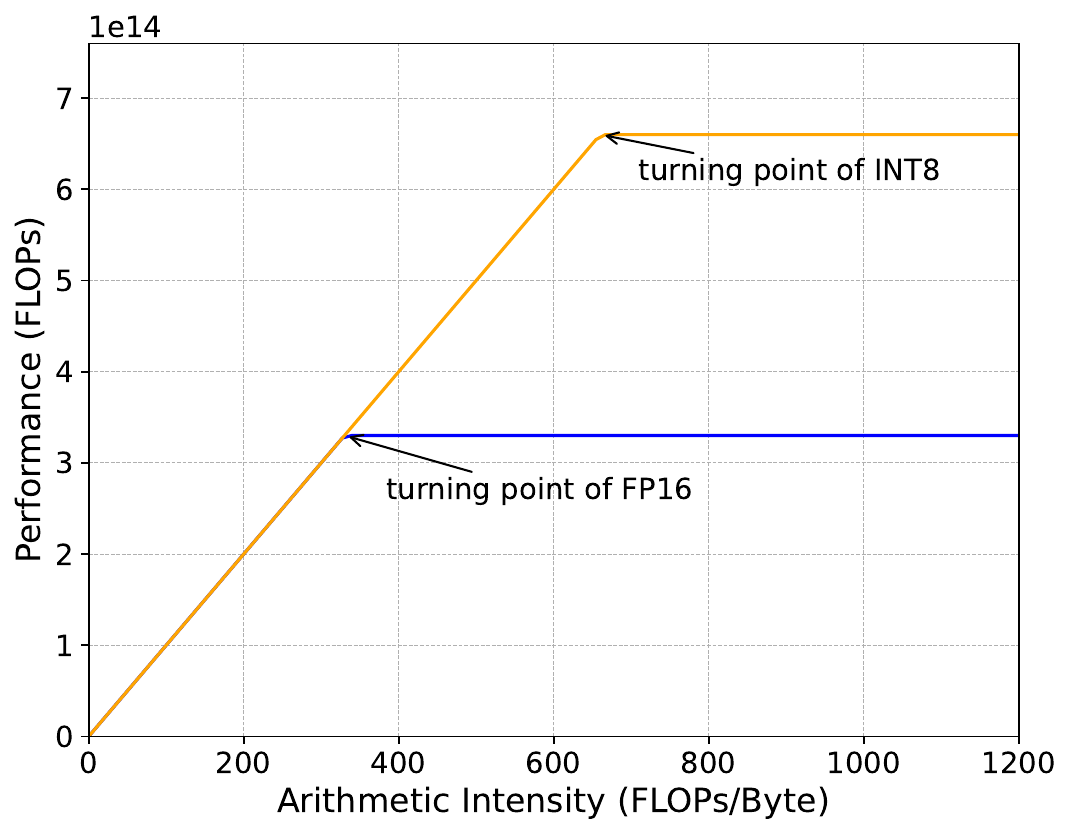}  
    \caption{The Roofline model for Nvidia RTX4090 GPU, with computations done in FP16 and INT8.}
    \label{fig:roofline}
\end{figure}

\subsection{\textbf{Roofline Model Analysis}} 
\label{subsec:roofline}

\begin{table}[t]
    \centering
    \renewcommand{\arraystretch}{1.5}
    \resizebox{0.5\textwidth}{!}{
    \begin{tabular}{cccc}
    \toprule
        Operations & FLOPs (B) & MOPs (B) & Arithmetic Intensity\\ \midrule \midrule
        qkv linear & 697M & 4.8M & 147\\
        qk matmul & 60M & 0.7M & 88\\ 
        sv matmul & 60M & 0.7M & 88\\ 
        FC1 & 929M & 6.2M & 149\\ 
        FC2 & 929M & 6.2M & 149\\ 
        softmax & 1.1M & 2.4M & 0.5\\ 
        layernorm & 0.9M & 0.61M & 1.5\\ 
        gelu & 2.0M & 0.61M & 3.25\\ \bottomrule
    \end{tabular}
    }
    \caption{Per-Operation FLOPs, memory operations (MOPs), and arithmetic intensity for the ViT-Base with an input image size of 224.}
    \label{tab:layers_roofline}
\end{table}

\begin{figure*}[ht]  
    \centering  
    \begin{subfigure}[t]{0.3\textwidth}  
        \includegraphics[width=\textwidth]{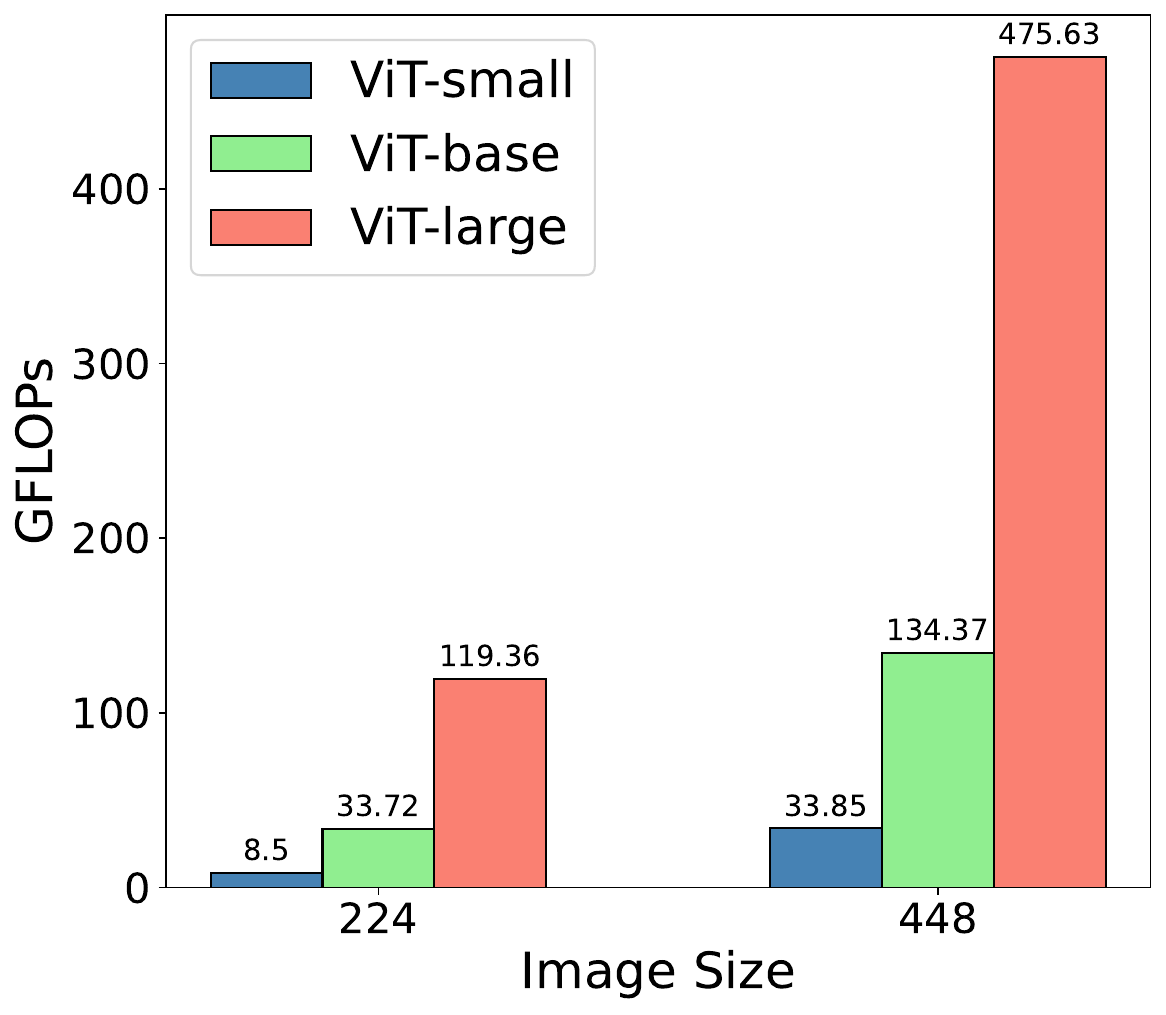} 
        \caption{GFLOPs}
        \label{fig:post_softmax}
    \end{subfigure}  
    \begin{subfigure}[t]{0.3\textwidth}  
        \includegraphics[width=\textwidth]{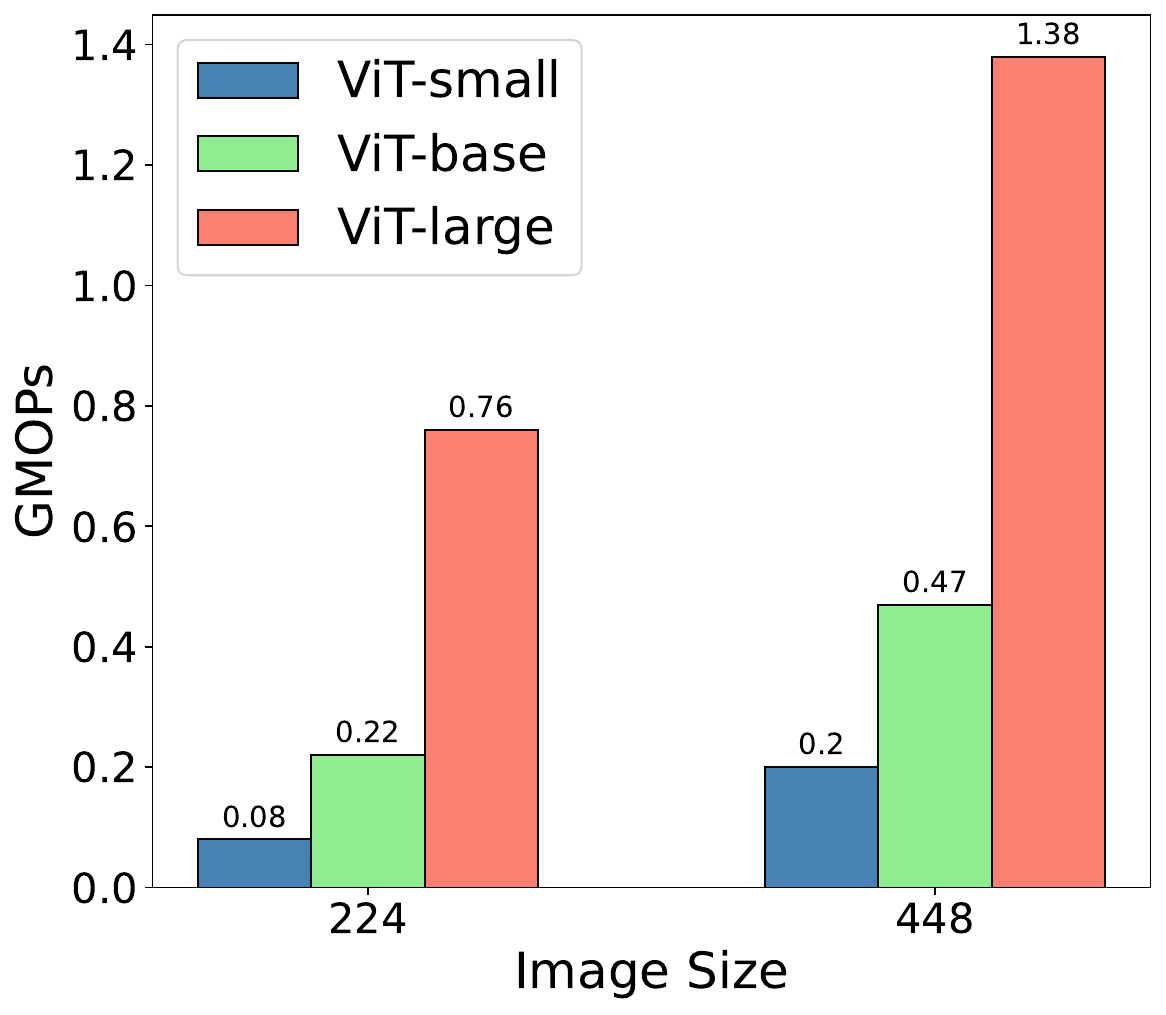}
        \caption{GMOPs}
        \label{fig:post_gelu}
    \end{subfigure} 
    \begin{subfigure}[t]{0.3\textwidth}  
        \includegraphics[width=\textwidth]{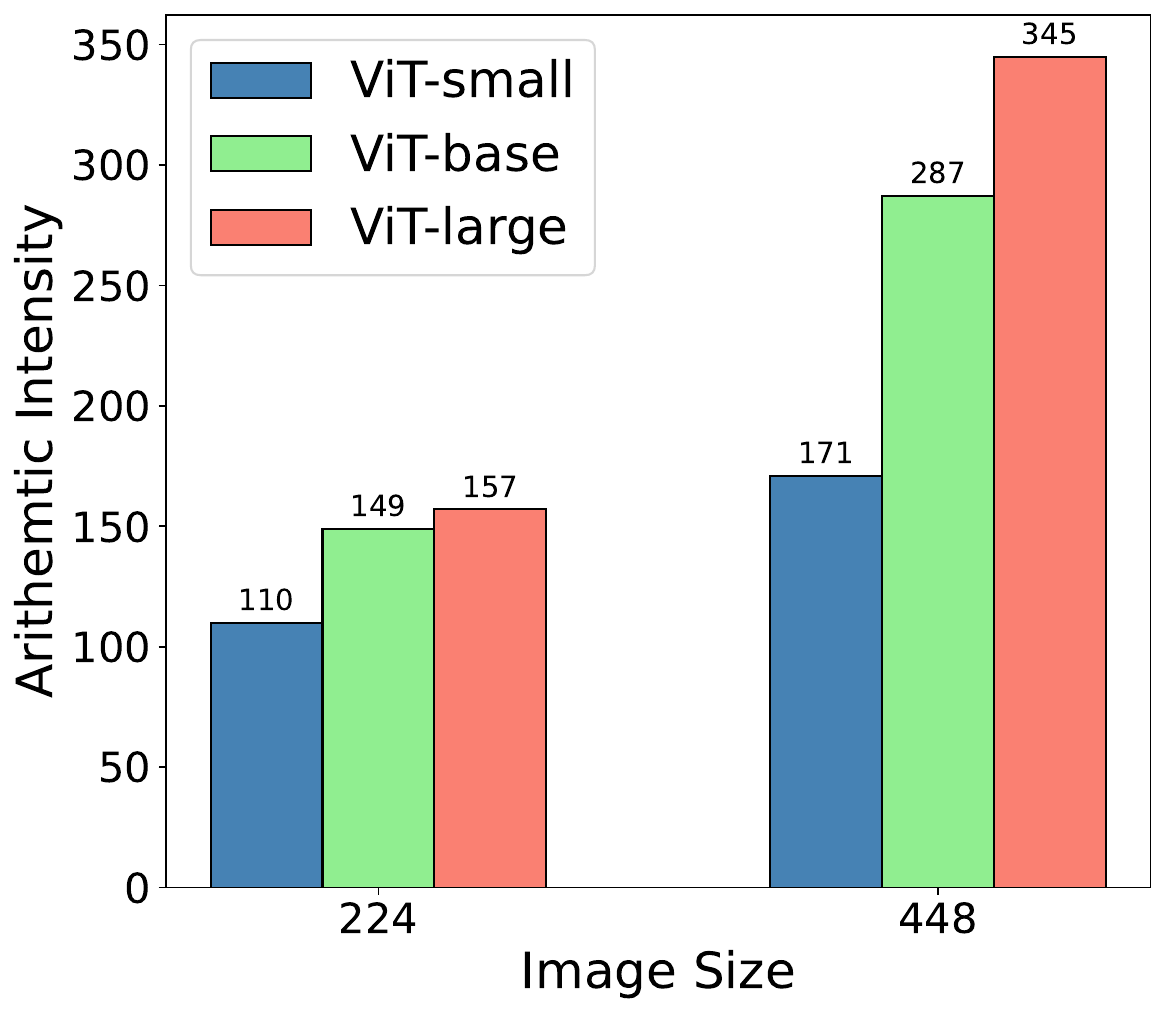}
        \caption{Arithemtic Intensity}
        \label{fig:post_layernorm}
    \end{subfigure} 
    \caption{This figure displays the GFLOPs, GMOPs, and Arithmetic Intensity for various sizes of ViTs with different image sizes.} 
    \label{fig:roofline_vits}
\end{figure*} 

The Roofline model \cite{roofline} is utilized to analyze ViTs. It provides a comprehensive framework for assessing efficiency when deployed on specific hardware (eg. RTX4090 GPU). This model aids in identifying whether a layer or operation is a computation or memory bottleneck, thereby enabling optimal utilization of memory access and processing capabilities \cite{roofline_llm}.

In alignment with the roofline model, we evaluate the model's computational demand by measuring both the floating-point operations (FLOPs) and the memory operations (MOPs) involved. Following this, we determine the arithmetic intensity, calculated as the ratio of FLOPs to bytes accessed (FLOPs / B), as depicted in Eq.\ref{eq:Arithmetic_I}.

\begin{equation}
\text{Arithmetic Intensity}=\frac{\#\mathrm{FLOPs}}{\#\mathrm{MOPs}}.
\label{eq:Arithmetic_I}
\end{equation}

In addition to arithmetic intensity, the hardware's peak performance, particularly with smaller bitwidth data, significantly impacts efficiency. For example, NVIDIA's RTX 4090 GPU doubles its speed from 330 TOP/s in FP16 to 660 TOP/s in INT8. As shown in Fig.\textcolor{red}{\ref{fig:roofline}}, the roofline model incorporates this by raising the performance ceiling when quantization is used, indicating better performance for compute-bound layers. 

\subsubsection{\textbf{Operations Analysis}} 
As illustrated in Table \ref{tab:layers_roofline}, we have analyzed the per-operation FLOPs, MOPs, and arithmetic intensity for the ViT-base, given an input image size of 224. The `FC1' and `FC2' layers showcase the highest arithmetic intensity, indicating efficient utilization of computing resources. In contrast, operations such as `qk matmul' and `sv matmul' exhibit lower arithmetic intensity due to the division of query, key, and value activations into smaller chunks (see eq.\ref{eq:head}), resulting in smaller matrix dimensions. The softmax operation, while requiring considerable memory access, uses fewer FLOPs, potentially posing an overhead during inference. 

When operating with advanced GPUs like the RTX 4090, operations with an arithmetic intensity below 200 are recognized as memory-bound, limiting their performance potential. In these instances, the focus should be on optimizing for faster memory access. Employing quantization to represent data in 8-bit or lower-bit formats can be especially advantageous, as it reduces the model's memory footprint, accelerates data transfer, thereby enhancing model inference.

\subsubsection{\textbf{End-to-end Analysis}} The analysis of ViTs across various image sizes, as demonstrated in Fig.\textcolor{red}{\ref{fig:roofline_vits}}, reveals that the computational demands, in terms of FLOPs and MOPs, increase more than proportionally with the size of the image. The underlying cause of this super-linear growth is the quadratic complexity associated with the activation multiplication in relation to the sequence length $l$. To illustrate, with a feature dimension $d$ and a number of heads $h$, the dimensionality of the `qk matmul' becomes $l \times d/h \times l$.

In particular, the ViT-large model shows increased arithmetic intensity, mainly attributable to its larger feature dimension $d$(where ViT-large has 1024 compared to ViT-small's 384). As the image size increases, this model tends to become compute-bound, particularly when using standard precision formats like FP16. Within the context of the roofline model on an RTX4090, as depicted in Fig.\textcolor{red}{\ref{fig:roofline}}, the performance of the model aligns with the blue line, signifying a compute-bound condition. Adopting INT8 precision emerges as a crucial optimization in such compute-bound situations. This approach not only mitigates computational bottlenecks but also capitalizes on the enhanced efficiency and throughput of quantized computing, significantly boosting overall performance.

\section{\textbf{Fundamental of Quantization}}
\label{sec:fund_quant}

In this section, we first introduce common quantization concepts \cite{gholami2022survey}. Afterward,  we discuss different quantization approaches including PTQ (Post-Training Quantization), QAT (Quantization-Aware Training), and DFQ (Data-Free Quantization). 

\subsection{\textbf{Linear Quantization}}
Linear Quantization linearly maps the continuous range of weight or activation values to a discrete set of levels. This process involves three key steps: scaling, rounding, and zero-point adjustment. As described by Jacob et al. \cite{jacob2018quantization}, the linear quantization function is as eq.\ref{eq:linear_quant}.
\begin{equation}
    q = round(\frac{r}{S}) + Z,
    \label{eq:linear_quant}
\end{equation}
where $r$ represents input real numbers, $q$ represents output quantized integers, $S$ is a real-valued scaling factor, and $Z$ is an integer zero point. This mapping is linear, meaning that the quantization levels are evenly spaced. The uniform spacing simplifies the computation and is particularly efficient in hardware that favors uniform arithmetic operations. The quantized integers can be converted back to floating-point representations as eq.\ref{eq:linear_dequant}.
\begin{equation}
    \tilde{r} = S (q - Z),
    \label{eq:linear_dequant}
\end{equation}
This operation is referred to as dequantization. However, the retrieved real numbers, $\tilde{r}$ , are not identical to the original values $r$ because of the inherent approximation introduced by the rounding operation.

\subsection{\textbf{Symmetric and Asymmetric Quantization}}
Scaling factor $S$ and zero point $Z$ in eq.\ref{eq:linear_quant} are two significant hyper parameters. The scaling factor segments a specific range of input real values into several divisions (see eq.\ref{eq:scaling_factor}).
\begin{equation}
     S=\frac{r_{max}-r_{min}}{q_{max}-q_{min}}
     \label{eq:scaling_factor}
\end{equation}
where $[r_{min}, r_{max}]$ and $[q_{min}, q_{max}]$ represent the clipping range of real and integer values, respectively. Calibration refers to the process of identifying the clipping range. If the scale factor $S$ is directly derived from $r_{min}$ and $r_{max}$, this method is termed asymmetric quantization, as the defined clipping range does not exhibit symmetry around the zero point.  Asymmetric quantization is more suitable when ranges are not symmetric or skewed. However, the symmetric quantization replaces the numerator of eq.\ref{eq:scaling_factor} with the maximum of absolute real values $r_{max}-r_{min}=2 \max(|r_{max}|,|r_{min}|)=2max(|r|)$. If the clipping range is symmetric, the value of zero point Z becomes 0. Eq.\ref{eq:scaling_factor} can be simplified as $q = round(r/S)$, which facilitates simpler implementation and more efficient inference. As for the $q_{min}$ and $q_{max}$, we can choose to use the “full range” or “restricted range”. In  “full range” mode, $ S=\frac{2max(|r|)}{2^{n}-1}$, where $n$ represents the quantization bit width and INT8 has the full range of $[-128, 127]$. In  “restricted range” mode, $ S=\frac{max(|r|)}{2^{n-1}-1}$ and INT8 has the full range of $[-127, 127]$. However, directly using the min/max values to determine the clipping range may be sensitive to outliers, resulting low resolution of quantization. Percentile \cite{NVIDIA2017TensorRT} and KL divergence \cite{wu2020integer} strategies are introduced to address this problem.

\subsection{\textbf{Static and Dynamic Quantization}}
The clipping range of weights can be computed statically before inference. However, the activations may vary. Depending on when the activations range is determined, we have dynamic quantization and static quantization. 

With dynamic quantization, the range is determined dynamically for every activation map during the runtime. This method necessitates the instantaneous calculation of input metrics (such as minimum, maximum, percentile, etc.), which can significantly increase computational costs. Despite this, dynamic quantization typically achieves greater precision because it precisely determines the signal range for each individual input.

With static quantization,  the range is pre-calculated and determined statically before the runtime. This method does not increase computational costs, though it often provides less accuracy. A common technique for pre-calculating the range is to process a set of calibration inputs to determine the average range of activations \cite{jacob2018quantization}. Mean Square Error (MSE) \cite{choukroun2019low} and entropy \cite{park2017weighted} are often used as metrics to choose the best range. Additionally, the clipping range can also be learned during the training process \cite{choi2018pact}.

\subsection{\textbf{Quantization Granularity}}
The determination of the clipping range can be categorized into layerwise, channelwise, and groupwise quantization, based on the level of granularity employed. As the granularity becomes finer, the computational overhead increases.

In layerwise quantization, the clipping range is determined by considering all the statistics of the entire parameters in a layer, and the same clipping range is used for all weights in that layer. This method is simple but may cause accuracy loss. For example, a weight matrix characterized by a more limited range of parameters may exhibit reduced quantization resolution due to the presence of other weight matrices that span a broader range.

Channelwise quantization is a common choice. In the same layer, different channels have more appropriate resolutions thanks to the use of separate scaling factors for corresponding channels. This often leads to improved accuracy.

In groupwise quantization, the clipping range is determined with respect to any groups of parameters in weights or activations. The granularity can be finer compared to layerwise and channelwise quantization. This method is helpful as the weights or activations may vary a lot within a single layer but this could also add considerable overhead.

\subsection{\textbf{Post Training Quantization}}
Post-Training Quantization (PTQ) is a technique that applies quantization operations to pre-trained models' weights and activations without the need for retraining or fine-tuning. It uses a pre-trained model and calibration data to calibrate and determine hyperparameters such as clipping ranges and scaling factors. After calibration, PTQ performs the quantization operation to produce a quantized model. Because PTQ requires only a small amount of data, it has a relatively low overhead. However, this may result in lower accuracy, especially for low-precision quantization.

\subsection{\textbf{Quantization Aware Training}}
Quantization-Aware Training (QAT) is a process that involves integrating quantization during training. Within this process, the backward pass, which includes gradient accumulation, is executed using floating-point calculations, and the high-precision weights are preserved during the entire retraining period. These weights will be quantized to integers during the forward pass. After retraining or fine-tuning the model with data, a quantized model can be obtained. Although QAT can recover the accuracy degradation caused by PTQ, it requires access to entire training data and is time-consuming, especially for extremely low bits like binary quantization.

\subsection{\textbf{Data Free Quantization}}
Data-Free Quantization (DFQ), alternatively termed Zero-Shot Quantization (ZSQ), operates quantization independently of actual data. This method ingeniously circumvents the conventional need for real-world datasets by synthesizing fake data that is similar to the original training data. The synthetic data thus generated is then utilized to calibrate the model during PTQ and to perform fine-tuning within QAT. The core advantage of DFQ lies in its capability to address concerns around data volume, privacy, and security. For instance, in domains where data is sensitive, such as healthcare or finance, DFQ offers a practical solution.
\section{\textbf{Model Quantization For ViTs}}
\label{sec:quant}

\begin{figure*}[ht]  
    \centering  
    \begin{subfigure}[t]{0.3\textwidth}  
        \includegraphics[width=\textwidth]{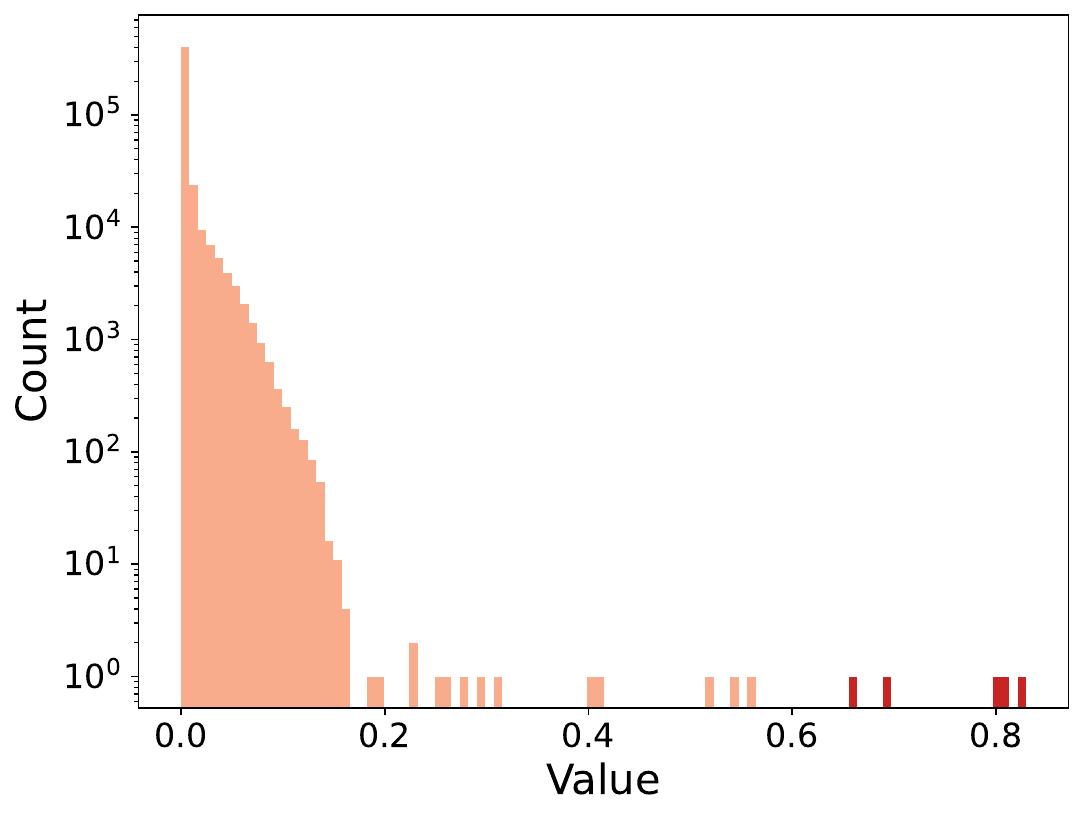} 
        \caption{Post-Softmax}
        \label{fig:post_softmax}
    \end{subfigure}  
    \begin{subfigure}[t]{0.3\textwidth}  
        \includegraphics[width=\textwidth]{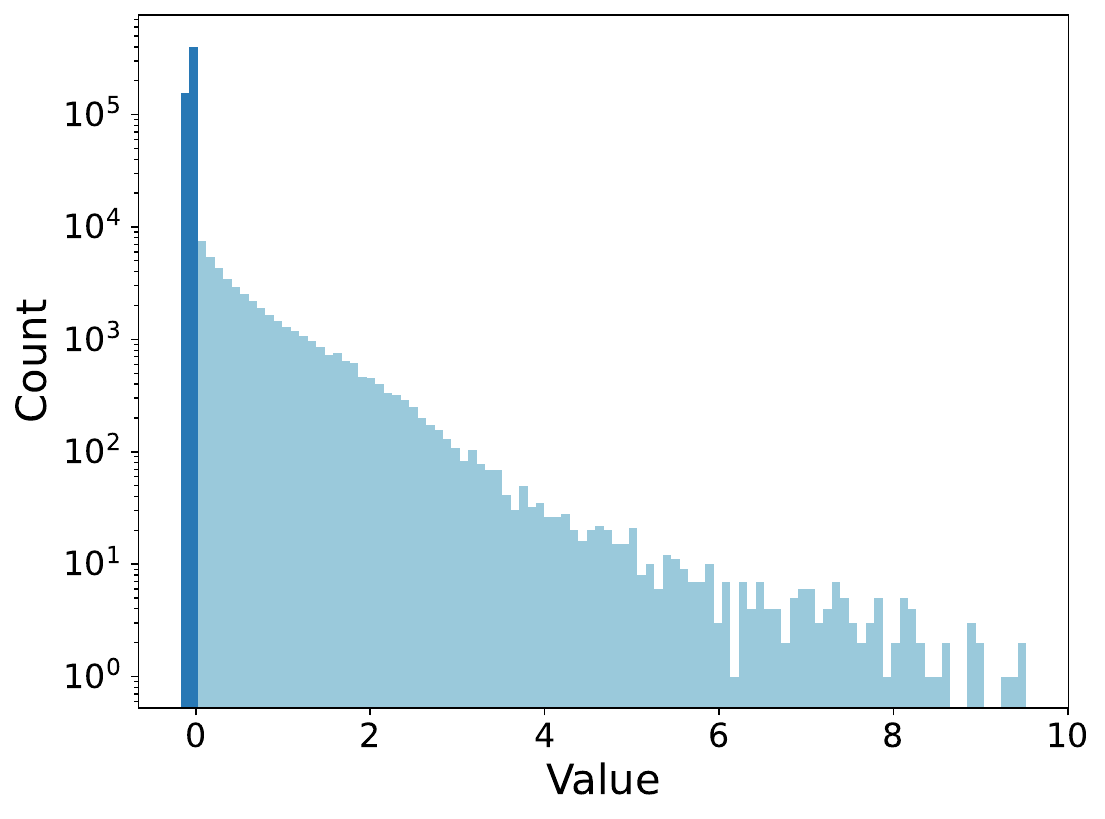}
        \caption{Post-Gelu}
        \label{fig:post_gelu}
    \end{subfigure} 
    \begin{subfigure}[t]{0.3\textwidth}  
        \includegraphics[width=\textwidth]{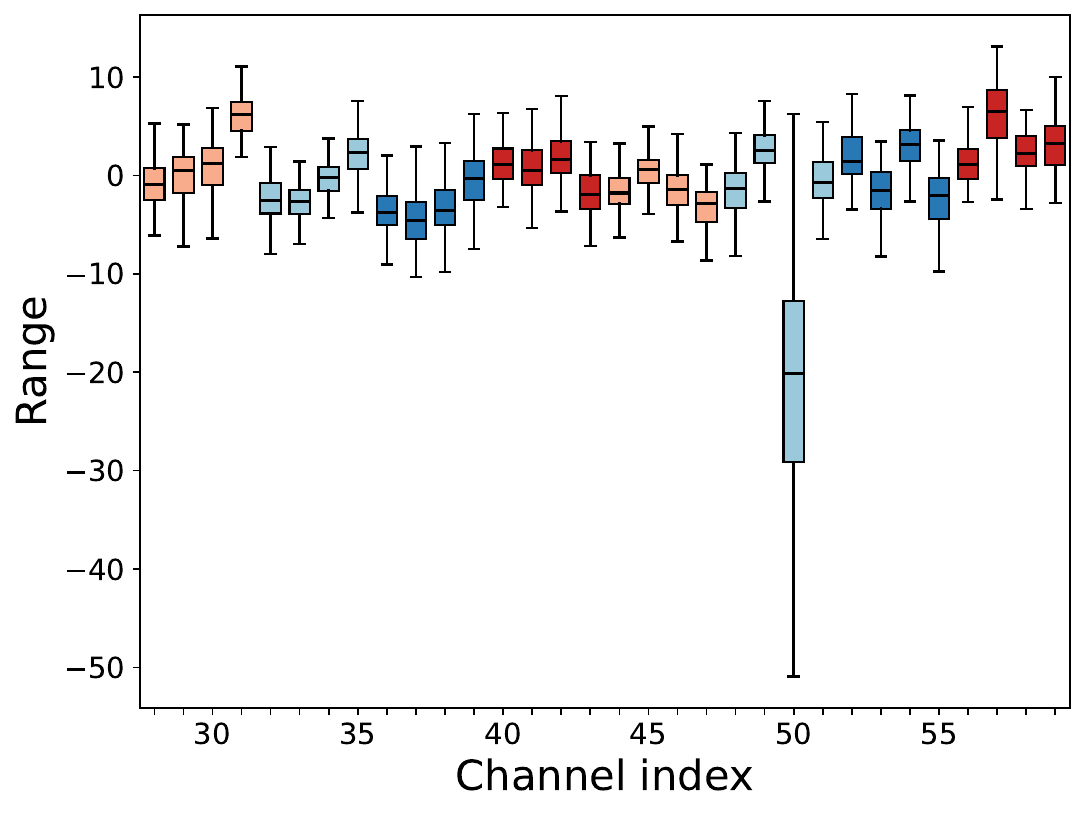}
        \caption{Post-LayerNorm}
        \label{fig:post_layernorm}
    \end{subfigure} 
    \caption{Distribution of the last module's post-Softmax, post-Gelu and Post-LayerNorm activation in Deit-Base.} 
    \label{fig:activation}
\end{figure*} 

This section critically evaluates research focused on enhancing the accuracy of quantized ViTs. It specifically addresses the challenges outlined in Sec.\textcolor{red}{\ref{subsection:activation_quantization}}, with providing solutions for the accuracy degradation issues identified during post-activation quantization of non-linear operations. Further, Sec.\textcolor{red}{\ref{subsec:calibration_ptq}} and \textcolor{red}{\ref{subsec:gradient_qat}} detail optimization techniques for PTQ and QAT respectively. Additionally, Sec.\textcolor{red}{\ref{subsec:Binary}} discusses tailored strategies for the complex task of binary quantization in ViTs.

\subsection{\textbf{Activation Quantization Optimization}} 
\label{subsection:activation_quantization}




As depicted in Fig.\textcolor{red}{\ref{fig:post_softmax}}, the activation distribution subsequent to the softmax function is highly unbalanced; a majority of values are concentrated near zero, while a minority (highlighted in darker hue) close to one. This presents a quantization challenge when using a single scaling factor, as a solitary outlier can disproportionately influence the uniform quantization precision across the entire tensor \cite{llm.int8}.

The histogram in Fig.\textcolor{red}{\ref{fig:post_gelu}} reveals a similar imbalance in value distribution post-GELU activation, with a preponderance of values clustered around zero. Moreover, the suitability for hardware-accommodating symmetric quantization is compromised due to the pronounced asymmetry of the distribution, particularly the limited range of negative values (indicated by the darker color).

Additionally, layer-wise quantization of Layernorm layer outputs leads to substantial performance decline, as evidenced by the pronounced inter-channel variability illustrated in Fig.\textcolor{red}{\ref{fig:post_layernorm}}. This variability underscores the challenge in maintaining uniform quantization efficacy across channels.

The following subsection examines various methodologies to address the above challenge when quantizing activations, corresponding to different operations such as Post-Softmax, Post-LayerNorm, and Post-GELU.

\subsubsection{\textbf{Post-Softmax Activation}} \label{subsubsec:post_softmax}
\begin{figure}[htbp]
    \centering
    \includegraphics[width=0.95\linewidth]{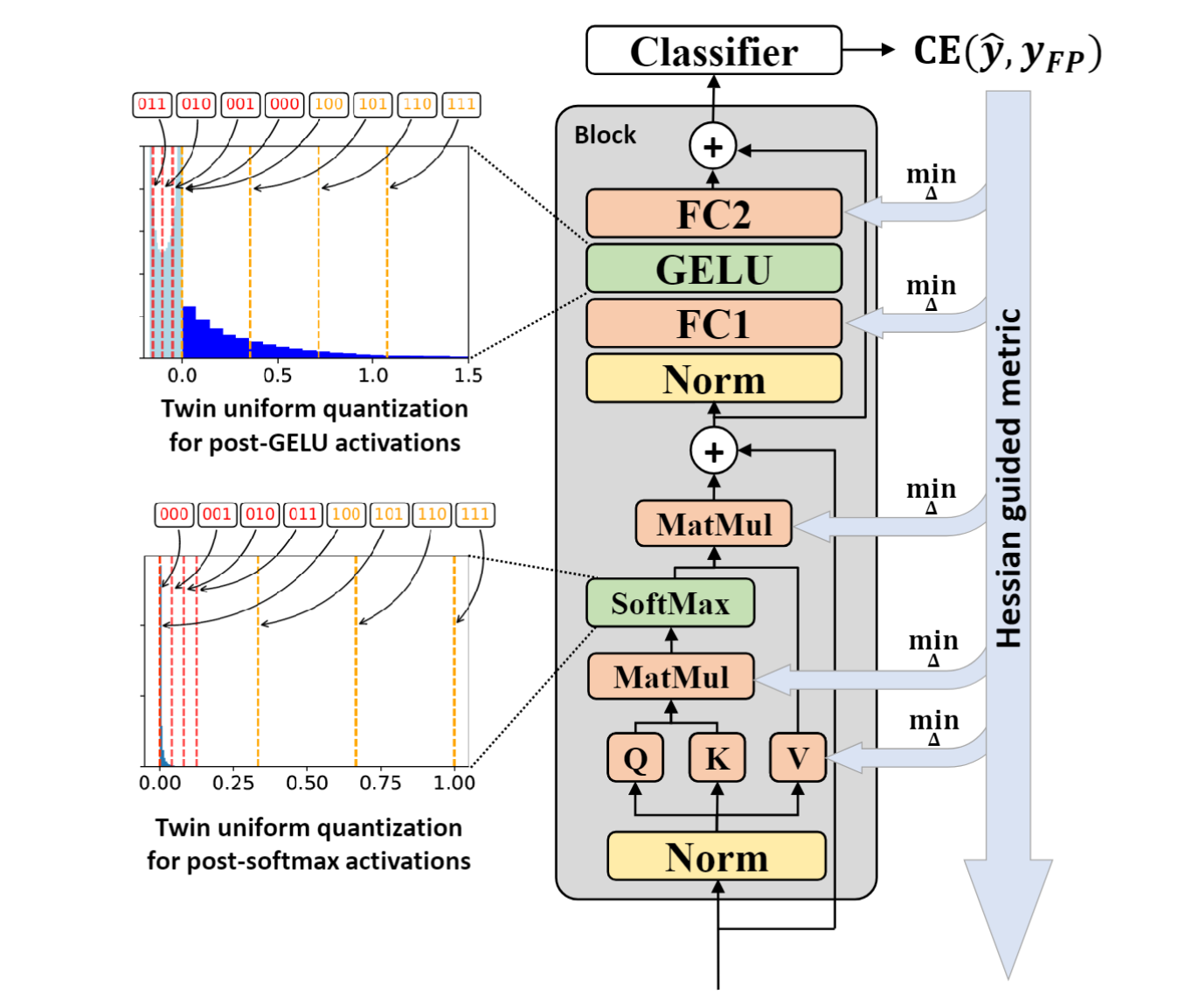}  
    \caption{Overview of the PTQ4ViT \cite{ptq4vit}.}
    \label{fig:ptq4vit}
\end{figure}

As shown in Fig.\textcolor{red}{\ref{fig:ptq4vit}}, PTQ4ViT \cite{ptq4vit} introduces the concept of twin uniform quantization for post-Softmax activations. This method divides the activation values following Softmax into two distinct quantization ranges, denoted as R1 and R2. Each range is governed by a unique scaling factor, $\Delta_{R1}$ for R1 and $\Delta_{R2}$ for R2, respectively. This dual-range strategy enables a more nuanced quantization process, effectively differentiating between smaller and larger activation values.

FQ-ViT \cite{fq-vit} adopts log2 quantization (see eq.\ref{eq:log2_quantization}) and assigns more bins to the frequently occurring small values found in post-softmax activation (attention maps), in contrast to the 4-bit uniform quantization, which allocates only one bin for these values.
\begin{equation}
Q(X|b) = \text{sign}(X) \cdot \text{clip}\left( \left\lfloor - \log_2 \frac{|X|}{\max(|X|)} \right\rceil, 0, 2^{b-1} - 1 \right).
\label{eq:log2_quantization}
\end{equation} 

APQ-ViT \cite{apq-vit} addresses the limitations of prior methods, which focused predominantly on mutual information while overlooking the Softmax function's Matthew effect. It implements an asymmetric linear quantization, tailored to maintain this effect during quantization, as shown in eq.\ref{eq:mpq}. The method hinges on a quantization step size proportional to the Softmax output's maximum value, promoting a more uniform distribution of quantization error across the Softmax range.
\begin{align}
\hat{x}_s &= \text{clamp}\left(\left\lfloor\frac{\text{softmax}(x)}{\Delta}\right\rceil, 0, 2^k - 1\right), \label{eq:mpq} \\
\Delta &= \frac{\max(\text{softmax}(x))}{2^k - 1} \notag.
\end{align}

RepQ-ViT \cite{repq-vit}, RepQuant \cite{repquant} and LRP-QViT \cite{lrp-qvit} initially employs $log\sqrt{2}$ quantization. This approach is better suited for the properties of attention scores in the activations with power-law distributions. Subsequently, the scales are reparameterized to change the base to 2, enabling bit-shifting operations in inference, which are more hardware-friendly.

TSPTQ-ViT \cite{teptq-vit} exploits the bit sparsity in non-normally distributed values and assigns different scaling factors to different regions of the post-softmax values.

I\&S-ViT \cite{IS-ViT} introduces the ``Shift-Uniform-Log2 Quantizer" (SULQ), which incorporates a shift bias prior to the log2 transformation. This approach allows for a comprehensive representation of the input domain's full range and effectively aligns with the long-tail distribution characteristic of post-Softmax activations.

\begin{table*}[ht]
    \centering
    \renewcommand{\arraystretch}{1.5} 
    \resizebox{1.0\textwidth}{!}{
    \begin{tabular}{c|c|ccc|ccc}
    \toprule
        \multirow{2}{*}{\textbf{Model}} & \multirow{2}{*}{\textbf{Retrain}} & \multicolumn{3}{c|}{\textbf{Activation}} & \multicolumn{3}{c}{\textbf{Accuracy (81.85 in FP32)}} \\ 
        ~ & ~ &Post-Softmax & Post-LayerNorm & Post-GELU & W8/A8 & W6/A6 & W4/A4 \\ \midrule \midrule
        PTQ4ViT \cite{ptq4vit} & \XSolid & \Checkmark & ~ & \Checkmark & 81.48 & 80.25 & - \\ 
        FQ-ViT \cite{fq-vit} & \XSolid & \Checkmark & \Checkmark & ~ & 81.20 & - & - \\ 
        APQ-ViT \cite{apq-vit} & \XSolid & \Checkmark & ~ & ~ & 81.72 & 80.42 & 67.48 \\ 
        RepQ-ViT \cite{repq-vit} & \XSolid & \Checkmark & \Checkmark & ~ & - & 81.27 & 75.61 \\ 
        RepQuant \cite{repquant} & \XSolid & \Checkmark & \Checkmark & ~ & - & 81.41 & 78.46 \\ 
        TSPTQ-ViT \cite{teptq-vit}  & \XSolid & \Checkmark & \Checkmark & \Checkmark & 81.72 & 80.25 & - \\ 
        I\&S-ViT \cite{IS-ViT} & \Checkmark & \Checkmark & \Checkmark & ~ & - & 81.68 & 79.97 \\ 
        MPTQ-ViT \cite{mptq} & \XSolid & Keep in FP & \Checkmark & \Checkmark & - & 81.25 & 76.14 \\
        LRP-QViT \cite{lrp-qvit} & \XSolid & \Checkmark & \Checkmark & ~ & - & 81.39 & 77.40 \\
        \bottomrule
    \end{tabular}
    }
    \caption{Top-1 Accuracy for Quantized \textbf{DeiT-Base} on ImageNet: The table details the Top-1 accuracy for models quantized with different post-activation optimizations and bit-width configurations. Baseline accuracy in FP32 is provided.}
    \label{tab:activation}
\end{table*}

\subsubsection{\textbf{Post-LayerNorm Activation}}
FQ-ViT \cite{fq-vit} proposes the Power-of-Two Factor (PTF) for Pre-LayerNorm quantization. The core idea of PTF is to assign different factors to different channels instead of different quantization parameters. This approach involves quantizing the input activation $X$ and applying layer-wise quantization parameters $s$ and $zp$, along with a Power-of-Two Factor $\alpha$ for each channel. The quantized activation $X_Q$ is then determined using the formula:
\begin{equation}
X_Q = \text{clip} \left( \frac{X}{2^{\alpha} s} + zp, 0, 2^b - 1 \right),
\end{equation}
where \( s \) and \( zp \) are calculated based on the maximum and minimum values of \( X \), and \( \alpha_c \) is chosen to minimize the quantization error for each channel \( c \).

RepQ-ViT \cite{repq-vit} proposes the Scale Reparam method for Post-LayerNorm activations involving a process that begins with applying channel-wise quantization to accurately capture their severe inter-channel variations. This initial step is followed by a critical reparameterization: transforming the channel-wise quantization scales and zero points into a more hardware-friendly layer-wise format. In accordance with this transformation, the approach additionally modifies the affine factors of LayerNorm and the weights of the following layer. Extending this concept, RepQuant \cite{repquant} advocates for a per-channel dual clipping strategy that establishes specific numerical upper and lower limits for each channel, aiming for accurate quantization while minimizing bias within the quantization space.

TSPTQ-ViT \cite{teptq-vit} introduces a solution by identifying outlier channels in the data and assigning them a unique scaling factor, distinct from the scaling factors of other channels. This process involves using the K-means algorithm to detect outliers based on the absolute maximum values across all channels. After identifying the outliers, the Hessian-guided scaling factors are determined from a set of linearly divided candidates. These scaling factors are then co-optimized with the scaling factors of weight for effective data processing.

I\&S-ViT \cite{IS-ViT} introduces ``smooth optimization strategy" (SOS), which contains s three stages to handle the high-variant activations.
During the first stage, the model is fine-tuned with full-precision weights and activations after LayerNorm that are quantized on a per-channel basis, while the rest of the activations are quantized on a per-layer basis. In the following stage, the model shifts from channel-wise to layer-wise quantization by employing the scale reparameterization technique. The final stage involves additional fine-tuning with both weights and activations quantized, to restore performance degradation caused by weight quantization.

MPTQ-ViT \cite{mptq} follows Smoothquant \cite{smoothquant} to operate equivalent transformation on the post-LayerNorm value and introduces a bias term to shift the distribution, making it more symmetric around zero.

LRP-QViT \cite{lrp-qvit} proposes clipped reparameterization, which involves clipping outliers within each channel's activations to mitigate variations. This is achieved by adjusting the scale and zero-point parameters obtained from channel-wise quantization, followed by reparameterizing the LayerNorm's affine factors and the subsequent layer's weights and biases to compensate for the distribution shifts induced by clipping. 

\subsubsection{\textbf{Post-GELU Activation}}
PTQ4ViT \cite{ptq4vit} applies twin uniform quantization to post-GELU activations, akin to its implementation in post-Softmax scenarios (refer to Sec.\textcolor{red}{\ref{subsubsec:post_softmax}}). In this context, the quantization process differentiates activation values based on their sign: negative values fall within range R1 and are quantized using a more refined scaling factor ($\Delta_{R1}$), whereas positive values are categorized under range R2 and quantized with a comparatively larger scaling factor ($\Delta_{R2}$).

In comparison to its application for post-Softmax values, TSPTQ-ViT \cite{teptq-vit} incorporates a similar two-region strategy with a distinct adaptation for the presence of a sign bit. 

While the traditional hand-crafted quantizers struggle to accurately represent post-GELU distributions, MPTQ-ViT \cite{mptq} proposes a data-dependent approach to automatically determine region-specific scaling factors, thereby eliminating the need for manual calibration and enhancing adaptability. The method involves dividing the post-GeLU values into three distinct regions based on their magnitude: negative values, small positive values, and large positive values. Each region is then assigned a specific SF, with the method ensuring hardware-friendly alignment by adjusting the bit shifts.

\textbf{Summary and Comparative analysis:} In reviewing the optimization strategies for activation quantization outlined in Table \ref{tab:activation}, several insights emerge. Notably, every quantization technique incorporated optimization after the softmax activation, underscoring its critical role and sensitivity in quantization processes. Interestingly, the GELU activation does not significantly influence quantization, suggesting a level of robustness. Moreover, all methods manage to achieve near-original accuracy when quantizing to 8-bit, showcasing their innovative designs tailored to the unique aspects of ViTs architectures.

When the bit depth is reduced to 4-bit, I\&S-ViT \cite{IS-ViT} achieves the most superior performance, which may be attributed to the utilization of a log2 quantizer adept at handling the unbalanced distribution following the softmax output. This strategy, coupled with a feasible fine-tuning process on a limited dataset, proves to be effective, indicating that strategic quantizer design and targeted fine-tuning can yield substantial benefits even at lower bit-widths.

\subsection{\textbf{Calibration Optimization For PTQ}} \label{subsec:calibration_ptq}

PTQ emerges as a compelling alternative to QAT by offering significant savings in GPU resources, time, and data requirements. The essence of PTQ lies in identifying an optimal quantization scale factor, which is contingent upon minimizing the discrepancy between the full-precision model and its quantized counterpart. This process is pivotal in preserving the fidelity of the model's output post-quantization. In this subsection, we review various methodologies that have been proposed to define and utilize different optimization distances during the calibration process, aiming to ascertain the most suitable quantization parameters for a quantized ViTs.

\subsubsection{\textbf{Pearson Correlation Coefficient and Ranking Loss}}
\begin{figure}[htbp]
    \centering
    \includegraphics[width=0.95\linewidth]{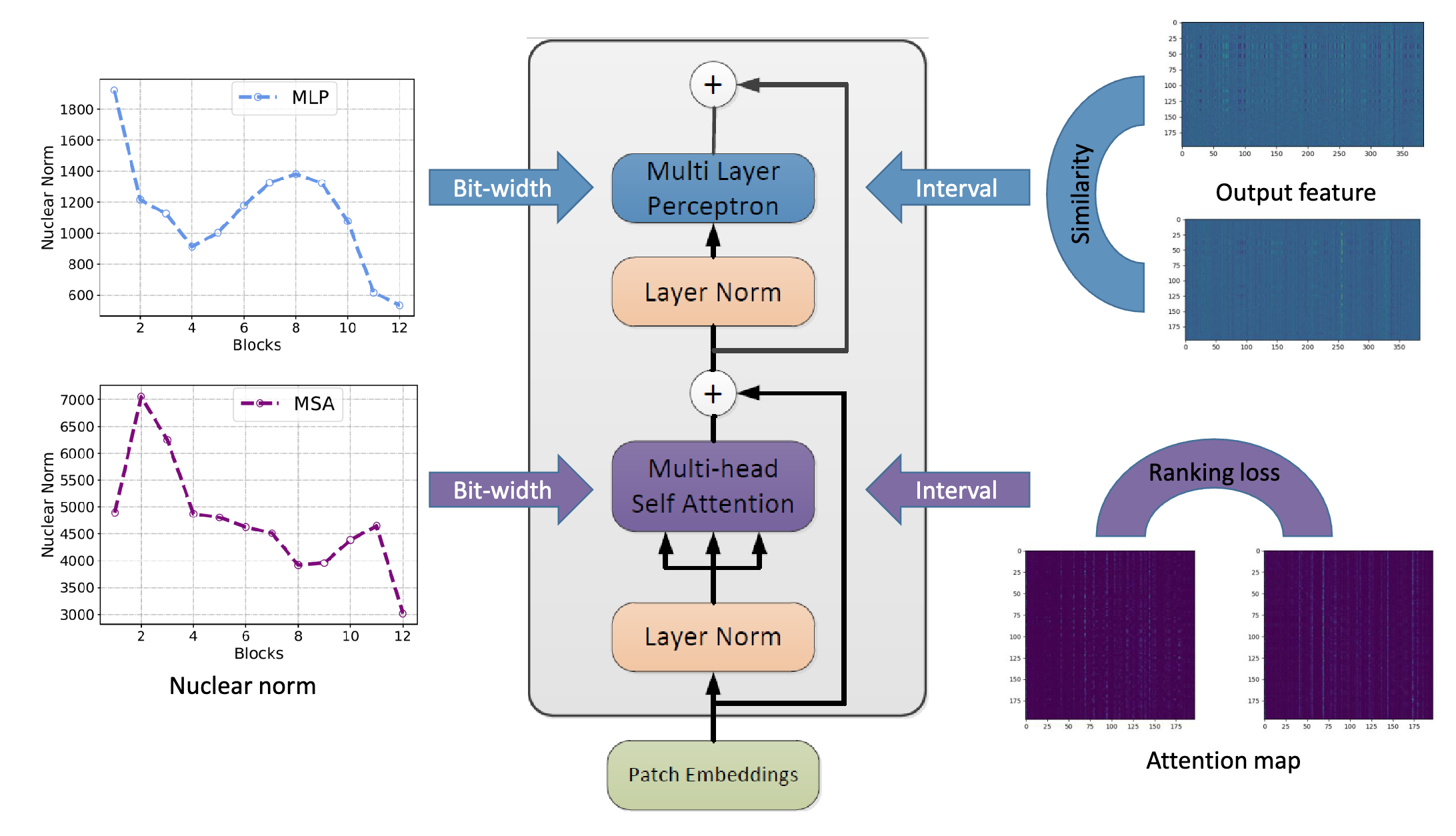}  
    \caption{The illustration presents similarity-aware quantization strategies applied to linear operations and ranking-aware quantization strategies for self-attention layers. These approaches are designed to enhance the quantization range efficiency. \cite{liu_ptq}.}
    \label{fig:liu_ptq}
\end{figure}

Liu et al. \cite{liu_ptq} utilize the Pearson Correlation Coefficient to assess the congruence between the original and quantized outputs within transformer modules. To address the issue of quantization affecting the order of attention scores, which is crucial for performance, they introduce a ranking loss. This loss is specifically formulated to preserve the self-attention layer's output hierarchy, thereby maintaining the transformer's unique capability to discern global feature relevance. The calibration process focuses on fine-tuning the quantization intervals for weights and inputs, leveraging a smaller calibration dataset to enhance the similarity between the original and quantized outputs, as elucidated in Fig.\textcolor{red}{\ref{fig:liu_ptq}}.

\subsubsection{\textbf{Hessian Guided Metric}}
PTQ4ViT \cite{ptq4vit} highlights the efficacy of a Hessian-guided metric for determining scaling factors, which can notably enhance the performance of quantized vision transformers, in contrast to other metrics that may not align with task-specific losses. The quantization-induced perturbations are quantifiable via a Taylor series expansion, as outlined in eq.\ref{eq:taylor}. To bypass the extensive computational demands of the full Hessian matrix, which requires second-order gradient computations, an approximate optimization strategy is employed, detailed in eq.\ref{eq:hessian}. Building upon this, APQ-ViT \cite{apq-vit} observes that in ultra low-bit scenarios, the Hessian Guided Metric might overlook crucial errors. To counteract this, the approach is refined to account for quantization errors in a blockwise fashion, allowing for a more nuanced perception of errors across adjacent layers. Additionally, a bottom-elimination mechanism is implemented to prioritize errors that substantially affect the final model output, rather than considering the entire error landscape.
\begin{equation}
    \mathbb{E}[L(\hat{W})]-\mathbb{E}[L(W)]\approx\epsilon^T\bar{g}^{(W)}+\frac12\epsilon^T\bar{H}^{(W)}\epsilon.
    \label{eq:taylor}
\end{equation}
\begin{equation}
    \min_\Delta\mathbb{E}[(\hat{O^l}-O^l)^Tdiag\big((\frac{\partial L}{\partial O_1^l})^2,\ldots,(\frac{\partial L}{\partial O_{|O^l|}^l})^2\big)(\hat{O^l}-O^l)].
    \label{eq:hessian}
\end{equation}

\subsubsection{\textbf{Layer-by-layer Reconstruction error}}
Evol-Q \cite{evol-q} employs a contrastive loss mechanism, drawing inspiration from self-supervised learning paradigms, to determine optimal quantization scales. This method utilizes a block-wise evolutionary search strategy to iteratively identify the best scales for each layer. The process involves completing the search for one block before proceeding sequentially to the next, ensuring a systematic and targeted approach to scale optimization.

With a simple layerwise squared loss, COMQ \cite{comq} utilizes coordinate descent optimization, where the scaling factors and bit-codes for each layer are sequentially optimized. This leads to backpropagation-free iterations that primarily involve dot products
and rounding operations without using any hyperparameters.


\subsubsection{\textbf{Quantization Error with Noisy Bias}}
The approach to quantization varies, with a linear quantizer using cosine similarity as suggested by EasyQuant \cite{easyquant}, and a non-linear one guided by the Hessian metric as proposed by PTQ4ViT \cite{ptq4vit}. In a novel contribution, NoisyQuant \cite{noisyquant} introduces a quantizer-agnostic method that enhances the quantization process by incorporating a Noisy Bias into the input activations prior to quantization. Specifically, a Noisy Bias $N$ is drawn from a uniform distribution $N\sim\mathcal{U}(-n,n)$, where the bounds are determined by $x\leq n\leq2b-x$. The search for an optimal $n$ employs a quantization error function, which is optimized using a subset of calibration data as defined in eq.\ref{eq:noisyloss} and eq.\ref{eq:noisyerror}.
\begin{equation}
    \mathcal{L}(n)=\sum_{x\in X}[D(x,N)].
    \label{eq:noisyloss}
\end{equation}

\begin{equation}
    \begin{aligned}&D(X,N)=QE(X+N)-QE(X)=\\&||(Q(X+N)-X-N)||_2^2–||(Q(X)-X)||_2^2\end{aligned}.
    \label{eq:noisyerror}
\end{equation}

\subsubsection{\textbf{Data-Free Calibration}}
\begin{figure}[htbp]
    \centering
    \includegraphics[width=0.9\linewidth]{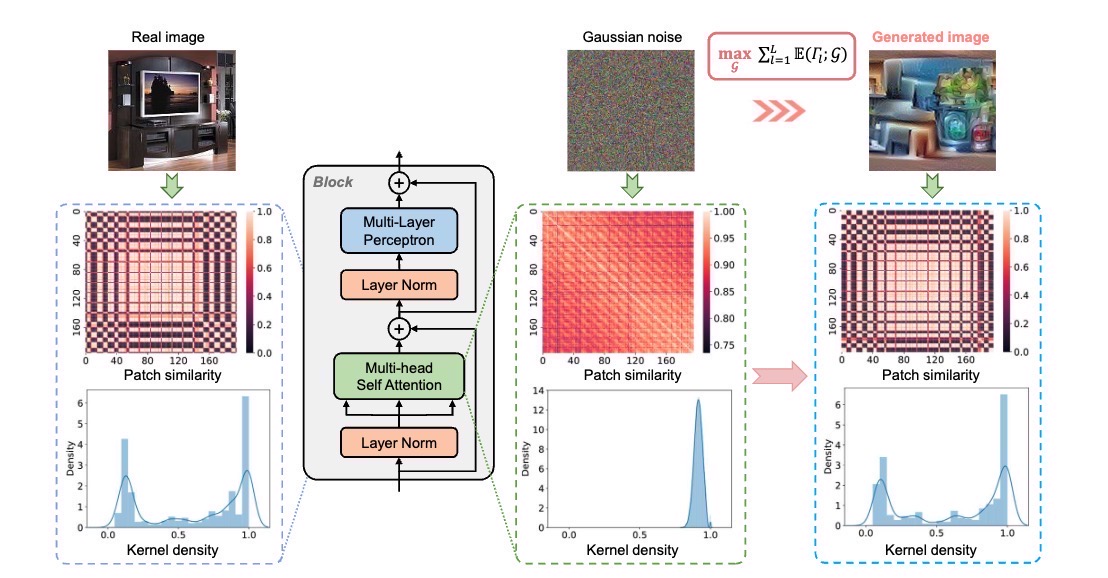}  
    \caption{Depiction of the approach for generating samples that are aware of patch similarity \cite{psaq}.}
    \label{fig:psaq-vit}
\end{figure}

PSAQ-ViT \cite{psaq} introduced an innovative Data-Free Quantization approach for ViTs, which eschews the need for real images during calibration. Leveraging the inherent diversity in patch similarity within the self-attention module of ViTs, PSAQ-ViT utilizes the differential entropy of this similarity as an objective to guide the optimization of Gaussian noise. This process aims to approximate the distribution of real images, as depicted in Fig.\textcolor{red}{\ref{fig:psaq-vit}}. Enhancing this methodology, PSAQ-ViT V2 \cite{psaq-vit-v2} incorporates an adaptive teacher-student paradigm, where synthetic samples are generated under the guidance of the full-precision (teacher) model. This is achieved through a minmax game designed to minimize model discrepancies. Significantly, this version removes the dependency on auxiliary category guidance in favor of task- and model-independent priors, broadening its applicability across various vision tasks and models.

\subsubsection{\textbf{Floating Point Format}}
Lee et al. \cite{fonf} present a comprehensive analysis of error models for both fixed-point and floating-point formats, aimed at identifying the most suitable numerical format during the calibration process. 

Further delving into the realm of floating-point quantization, LLM-FP4 \cite{llm-fp4} elaborates on the formulation of floating-point variables and their quantization. The study highlights the significance of carefully selecting exponent bits and scaling parameters to ensure efficient quantization outcomes. A search-based algorithm is introduced by the researchers to determine the optimal format and clipping range for each layer, specifically addressing the unique challenges associated with floating-point quantization. Additionally, LLM-FP4 proposes an innovative pre-shifted exponent bias method. This method is particularly effective in managing the high inter-channel variance often encountered in transformer models, thereby improving the precision of activations quantization.

\textbf{Summary and Comparative analysis:} The comparative results showcased in Table \ref{tab:acc_compare2} reveal significant insights into the performance of PTQ. It is apparent that most PTQ methods are optimized for 8-bit quantization, suggesting that quantizing to less than 8 bits could result in a loss of accuracy. The application of a layer-by-layer reconstruction error during calibration is shown to be effective. Yet, this technique necessitates an extended calibration duration to pinpoint suitable quantization scaling factors.

Additionally, LLM-FP4 \cite{llm-fp4} yields encouraging outcomes with a limited number of exponent and mantissa bits. This highlights the practicality of implementing floating-point formats for quantization at reduced bit-widths, indicating the prospect of achieving significant efficiency improvements on both existing and forthcoming hardware architectures.

\subsection{\textbf{Gradient-base Optimization For QAT}} \label{subsec:gradient_qat}
QAT has emerged as an effective approach in model quantization, particularly in addressing the constraints of PTQ. In contrast to PTQ, which determines quantized parameters from pre-trained full-precision models often leading to suboptimal performance, QAT seamlessly integrates quantization within the training cycle. This approach proves especially beneficial when scaling down to ultra-low bit precision, such as 4 bits or lower, where PTQ tends to struggle with significant performance loss.

QAT effectively reduces the performance degradation commonly associated with quantization by incorporating it into the backpropagation process. However, QAT is not without challenges. Key issues include the difficulty in approximating gradients for the non-differentiable quantization function and the occurrence of quantization oscillation, which can impede the optimization process.

This section delves into various strategies developed to enhance both the efficiency and effectiveness of quantized ViTs through innovative gradient-based optimization techniques.



\subsubsection{\textbf{Differentiable Search for Group Assignment}}
Quantformer \cite{quantformer} leverages entropy information as a capacity-aware metric to maintain the consistency of self-attention ranks by minimizing the discrepancy between quantized and full-precision self-attention layers. This method effectively upholds the rank order of self-attention with minimal computational costs. Furthermore, \cite{quantformer} introduces a differentiable search mechanism aimed at optimally grouping patch feature dimensions. This ensures that patch features within the same group adopt a unified quantization approach, using common thresholds and quantization levels. As a result, this strategy mitigates rounding and clipping inaccuracies across variably distributed patch features.

\subsubsection{\textbf{Distillation-assisted Training}}
Knowledge distillation (KD) \cite{distilling} is a prominent methodology within the transfer learning domain, where knowledge is transferred from a more extensive ``teacher" model to a more compact ``student" model. This approach is particularly effective in mitigating the accuracy loss encountered in models that have been compressed via QAT \cite{ternarybert, du2024bitdistiller}. 

Q-ViT \cite{q-vit-acc} follow \cite{deit} incorporates a distillation token for direct supervision of the classification output in the quantized ViT framework. To counteract the altered distribution in quantized attention modules, the Information Rectification Module (IRM) is introduced. The IRM, operating in the forward process, strategically maximizes information entropy to refine these distributions. Concurrently, the backward process employs a Distribution Guided Distillation (DGD) strategy. This approach focuses on reducing discrepancies in distribution by employing an attention similarity loss, effectively bridging the gap between the quantized ViTs and its full-precision analogue.

GPUSQ-ViT \cite{boostvit} utilize the calibration losses for the hard label, soft logits and feature maps with corresponding weight factors.

Huang et al. \cite{vavtq} present a new quantization technique that is aware of variations. This approach incorporates a multi-crop knowledge distillation strategy, which aids in stabilizing the training process and lessening the effects of variations encountered during QAT. Additionally, \cite{vavtq} proposes a module-dependent quantization scheme that dynamically adjusts the quantization process for each module, based on its specific characteristics. Moreover, to address the issue of weight oscillation during training, which can lead to further instability, the paper presents an Oscillation-aware Bin Regularization technique. This approach aims to stabilize the weight distribution and suppress oscillation, thereby enhancing the overall training process.

\subsubsection{\textbf{Progressive Training}}
Quantizing real-valued pretrained models to extremely low-bit representations poses significant challenges due to the cluttered and non-convex nature of the loss landscape. TerViT \cite{tervit} introduces an innovative approach where an 8-bit model is initially quantized and trained, followed by a transfer to ternary weight training. Building upon this, Bit-shrinking \cite{bitshrinking} introduces a novel technique that involves controlling a 'sharpness' term closely associated with the noise introduced during quantization. This method effectively smoothens the loss landscape, which is crucial for maintaining the accuracy of the model. This smoother transition is achieved by progressively reducing the bit-width of the model while simultaneously regulating any increase in sharpness. Such a strategy ensures the preservation of model accuracy throughout the quantization process.

\subsubsection{\textbf{Outlier-Aware Training}}
PackQViT \cite{packqvit} introduces an outlier-aware training approach, differing from the method in \cite{fq-vit} which calculates the power-of-two factor during calibration. In each training iteration of PackQViT, the process begins by searching for the channel index and the power-of-two coefficient of each outlier using $l_2$ minimization. Following this, it updates the quantized parameters to mitigate the adverse effects of outliers.

\begin{table*}[!ht]
    \centering
    \renewcommand{\arraystretch}{1.5} 
    \resizebox{1.0\textwidth}{!}{
    \begin{tabular}{c|c|c|cccc}
    \toprule
        \multirow{2}{*}{\textbf{Optimization}} & \multirow{2}{*}{\textbf{Method}} & \multirow{2}{*}{\textbf{Model}} & \multicolumn{4}{c}{\textbf{Accuracy (81.85 in FP32)}} \\ 
        ~ & - & - & W8/A8 & W4/A4 & W2/A2 & W1/A1 \\ \midrule \midrule
        \multirow{9}{*}{\textbf{Calibration Optimization For PTQ}} & Pearson Correlation Coefficient and Ranking Loss & Liu et al. \cite{liu_ptq} & 80.48 & - & - & - \\ \cline{2-7}
        ~ & \multirow{2}{*}{Hessian Guided Metric} & PTQ4ViT \cite{ptq4vit} & 81.48 & - & - & - \\ 
        ~ & ~ & APQ-ViT \cite{apq-vit} & 81.72 & 67.48 & - & - \\ \cline{2-7}
        ~ & \multirow{2}{*}{Layer-by-layer Reconstruction error} & Evol-Q \cite{evol-q} & 82.67 & - & - & - \\
        ~ & ~ & COMQ \cite{comq} & - & 78.72 & - & - \\ \cline{2-7}
        ~ & Quantization Error with Noisy Bias & NoisyQuant \cite{noisyquant} & 81.45 & - & - & - \\ \cline{2-7}
        ~ & \multirow{2}{*}{Data-Free Calibration} & PSAQ-ViT \cite{psaq} & 79.10 & - & - & - \\ 
        ~ & ~ & PSAQ-ViT V2 \cite{psaq-vit-v2} & 81.52 & - & - & - \\ \cline{2-7}
        ~ & \multirow{2}{*}{Floating Point Format} & Lee et al. \cite{fonf} & 81.88 & - & - & - \\ 
        ~ & ~ & LLM-FP4 \cite{llm-fp4} & - & 79.40 & - & - \\ \midrule \midrule
        \multirow{8}{*}{\textbf{Gradient-base Optimization For QAT}} & Differentiable Search for Group Assignment & Quantformer \cite{quantformer} & - & 79.70 & 73.80 & - \\ \cline{2-7}
        ~ & \multirow{3}{*}{Distillation-aware Training} & Q-ViT \cite{q-vit-acc}& - & 83.00 & 74.20 & - \\ 
        ~ & ~ & GPUSQ-ViT \cite{boostvit} & - & 81.60 & - & - \\ 
        ~ & ~ & Huang et al. \cite{vavtq} & - & 74.71\textsuperscript{†} & 59.73\textsuperscript{†} & - \\ \cline{2-7}
        ~ & \multirow{2}{*}{Progressive Training} & TerViT \cite{tervit}& - & - & 74.20 & - \\ 
        ~ & ~ & Bit-shrinking \cite{bitshrinking} & 81.56 & - & - & - \\ \cline{2-7}
        ~ & Outlier-Aware Training & PackQViT \cite{packqvit} & 82.90 & 81.50 & - & - \\ \cline{2-7}
        ~ & Oscillation-free Training & OFQ \cite{ofq} & - & 75.46\textsuperscript{†} & 64.33\textsuperscript{†} & - \\ \midrule \midrule
        \multirow{3}{*}{\textbf{Binary Quantization}} & Softmax-aware Binarization & BiViT \cite{bivit} & - & - & - & 69.6 \\ \cline{2-7}
        ~ & Gradient regularization Scheme and Activation Shift Module & Xiao et al. \cite{Xiao_binary} & - & - & - & 49.41\textsuperscript{†} \\ \cline{2-7}
        ~ & Integration With CNN & BinaryViT \cite{binaryvit} & - & - & - & 70.6 \\ \bottomrule
    \end{tabular}
    }
    \caption{Top-1 Accuracy for Quantized \textbf{DeiT-Base} on ImageNet (`†' signifies \textbf{DeiT-Tiny}, achieving 72.21 in FP32): The table details the Top-1 accuracy for models quantized with different optimizations and bit-width configurations. Baseline accuracy in FP32 is provided.}
    \label{tab:acc_compare2}
\end{table*}

\subsubsection{\textbf{Oscillation-free Training}}
QFQ \cite{ofq} investigates how weight oscillation in quantization-aware training negatively affects model performance, highlighting the role of the learnable scaling factor in exacerbating this issue. It propose three techniques to mitigate this problem: Statistical Weight Quantization (StatsQ) for robust quantization, Confidence-Guided Annealing (CGA) to stabilize oscillating weights, and Query-Key Reparametrization (QKR) to resolve intertwined oscillations in ViT's self-attention layers.

\textbf{Summary and Comparative analysis:} Comparison results shows in Table \ref{tab:acc_compare2}.
The comparative data presented in Table \ref{tab:acc_compare2} reveals distinct performance trends between PTQ and QAT methods. Notably, QAT methods demonstrate superior outcomes compared to PTQ, achieving lossless accuracy even when quantizing to extremely low bit-widths such as 4 bits, albeit with greater training costs. Distillation-aware training, despite its higher memory demands on hardware, exhibits the highest performance among the methods examined.

Moreover, the ability to maintain accuracy at 4 bits is stimulating interest in dedicated hardware accelerators capable of efficient computation at this bit-width. This reflects a growing trend towards optimizing computational resources without compromising the quality of results.

\subsection{\textbf{Binary Quantization}} \label{subsec:Binary}
Binary quantization in Vision Transformers (ViT) introduces ultra-compact 1-bit parameters, drastically reducing model size while enabling efficient bitwise operations. This method significantly cuts computational demands but faces challenges due to a considerable performance drop, primarily attributed to the stark reduction in parameter precision. This reduction hampers the model's ability to process complex information. This section delves into a range of innovative approaches developed to tackle these challenges, focusing on enhancing the binary quantization process to balance efficiency with performance retention.

BiViT \cite{bivit} proposes Softmax-aware Binarization that dynamically adjusts the binarization process, reducing the errors that typically arise when binarizing the softmax attention values. This allows for a more accurate representation of the attention mechanism

Xiao et al. \cite{Xiao_binary} includes a novel gradient regularization scheme (GRS) to reduce weight oscillation in binarization training and an activation shift module (ASM) to minimize information distortion in activation.

\begin{figure}[htbp]
    \centering
    \includegraphics[width=0.9\linewidth]{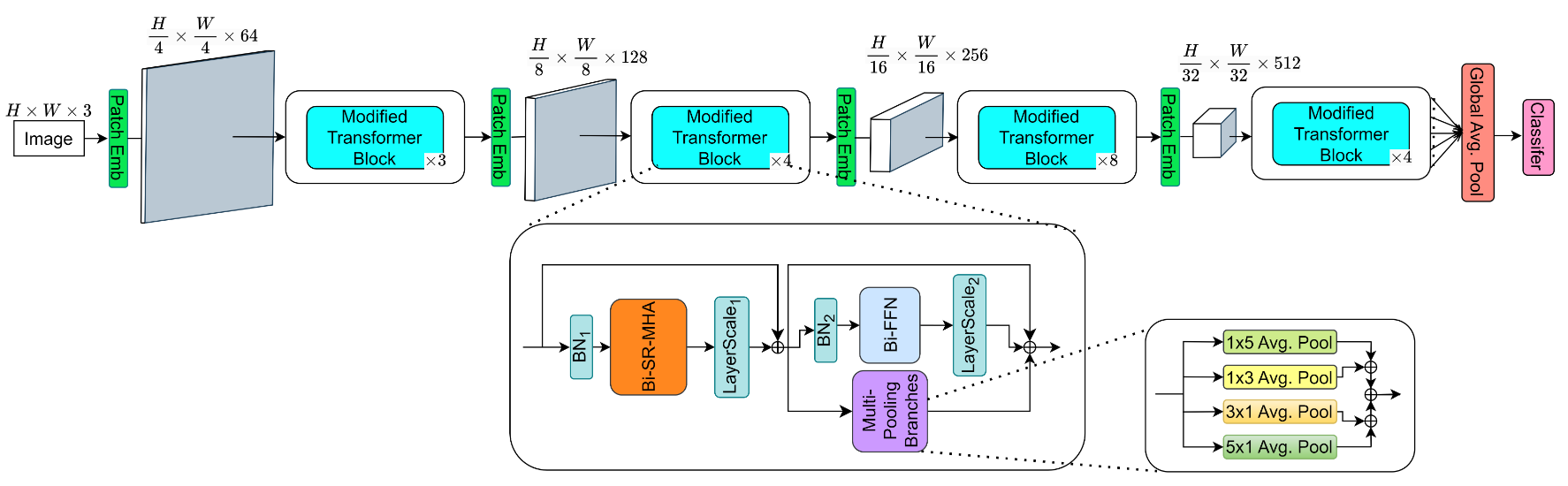}  
    \caption{Overview of BinaryViT's structure \cite{binaryvit}.}
    \label{fig:binaryvit}
\end{figure}

BinaryViT \cite{binaryvit} integrates key architectural features from CNNs into a pure ViTs architecture without introducing convolutions. Specifically, it integrates features such as global average pooling instead of cls-token pooling, multiple average pooling branches, affine transformations before residual connections, and a pyramid structure, as shown in Fig.\textcolor{red}{\ref{fig:binaryvit}}. These innovations aim to significantly increase the representational capability and computational efficiency of ViTs.

\textbf{Summary and Comparative analysis:} Examination of Table \ref{tab:acc_compare2} discloses a scarcity of Binary Quantization methodologies, indicative of the substantial challenges inherent in this domain.

The tabulated results underscore a pronounced accuracy deficit, surpassing a 10\% margin, relative to the full-precision model. This delineates the trade-off intrinsic to binary quantization: the diminution of computational overhead and storage is counterbalanced by a reduction in the model’s fidelity.

Nonetheless, current research trajectories are promising. Integrative approaches that amalgamate binary quantization with auxiliary modules or the formulation of novel architectures \cite{binaryvit, Xiao_binary}, have the potential to catalyze advancements. These ventures are directed towards reconciling the efficiency of binary quantization with the retention of robust model accuracy.

\section{\textbf{Hardware Acceleration For Quantized ViT}}
\label{sec:hardware}

Fully utilizing the advantages of Quantization, such as reduced memory consumption and increased computing capability, on a target hardware is critical. In this section, we discuss the restrictions imposed by non-linear operations and provide an overview of the optimizations in Sec.\textcolor{red}{\ref{sec:Non-linear}}. Following that, we introduce hardware accelerators with ViTs-specific considerations in Sec.\textcolor{red}{\ref{sec:accelerator}}.

\subsection{\textbf{Accelerating Non-Linear Operations}} \label{sec:Non-linear}

\begin{table}[b]
    \centering
    \resizebox{0.5\textwidth}{!}{
    \begin{tabular}{c|ccc|c|c}
    \toprule
        \textbf{Model} & \textbf{Softmax} & \textbf{LayerNorm} & \textbf{GELU} & \textbf{Retrain} & \textbf{W8A8 (81.85 in FP32)} \\ \midrule \midrule
        FQ-ViT \cite{fq-vit} & \Checkmark & \Checkmark & \XSolid & \XSolid & 81.20 \\ 
        PackQViT \cite{packqvit} & \Checkmark & \Checkmark & \Checkmark & \Checkmark & 82.90 \\ 
        I-ViT \cite{i-vit} & \Checkmark & \Checkmark & \Checkmark & \Checkmark & 81.74 \\ 
        EdgeKernel \cite{practical} & \Checkmark & \Checkmark & \Checkmark & \XSolid & 80.74 \\ 
        SOLE \cite{sole} & \Checkmark & \Checkmark & \Checkmark & \XSolid & 81.12 \\ \bottomrule
    \end{tabular}
    
    }
    \caption{Top-1 Accuracy of Quantized \textbf{DeiT-Base} on ImageNet: The table shows Top-1 accuracy for models with integer approximations of non-linear operations. Baseline accuracy in FP32 is provided.}
    \label{tab:non_linear_int}
\end{table}

Optimizing non-linear operations in the context of quantized ViTs is pivotal for several reasons. Firstly, low-bit computing units, which are central to quantization, offer significant benefits in terms of computational efficiency and memory footprint reduction. However, these units are predominantly optimized for linear operations and often lack native support for essential non-linear operations such as GELU, LayerNorm, and Softmax, which are integral to the ViTs architecture. This discrepancy poses a substantial challenge, as the absence of efficient non-linear computation mechanisms can negate the advantages offered by quantization, leading to bottlenecks in processing speed and energy efficiency.

Moreover, the reliance on 32-bit floating-point (FP32) arithmetic units for handling non-linear operations introduces significant overheads. A large portion of the inference time in ViT models is consumed by floating-point non-linear activations, normalizations, and the associated quantization and dequantization operations \cite{softermax}. This not only diminishes the throughput but also increases the energy consumption, thereby undermining the potential gains from quantization.

The main idea of optimizations is to replace floating-point nonlinear operations with integer approximations to unnecessary Quantization and dequantization such that they are both faster and cheaper to implement in specialized hardware accelerators.

As summarized in Table \ref{tab:non_linear_int}, we introduce optimization techniques in the subsequent section that aim to further enhance ViT inference efficiency.

FQ-ViT \cite{fq-vit} introduces "Log-Int-Softmax," an integer-only variant of the softmax function that approximates the exponential component using a second-order polynomial \cite{i-bert}. This method is coupled with Log2 quantization for efficient computation. For LayerNorm, FQ-ViT applies Powers-of-Two Scale factors to shift quantized activations, then computes the mean and variance using integer arithmetic, enhancing hardware compatibility.

PackQViT \cite{packqvit} eschews the second-order polynomial approximation in favor of a more straightforward approach that substitutes the natural constant
$e$ with 2 in the softmax equation. This simplification results in no observable loss in accuracy after retraining.

\begin{figure}[h]
    \centering
    \includegraphics[width=0.9\linewidth]{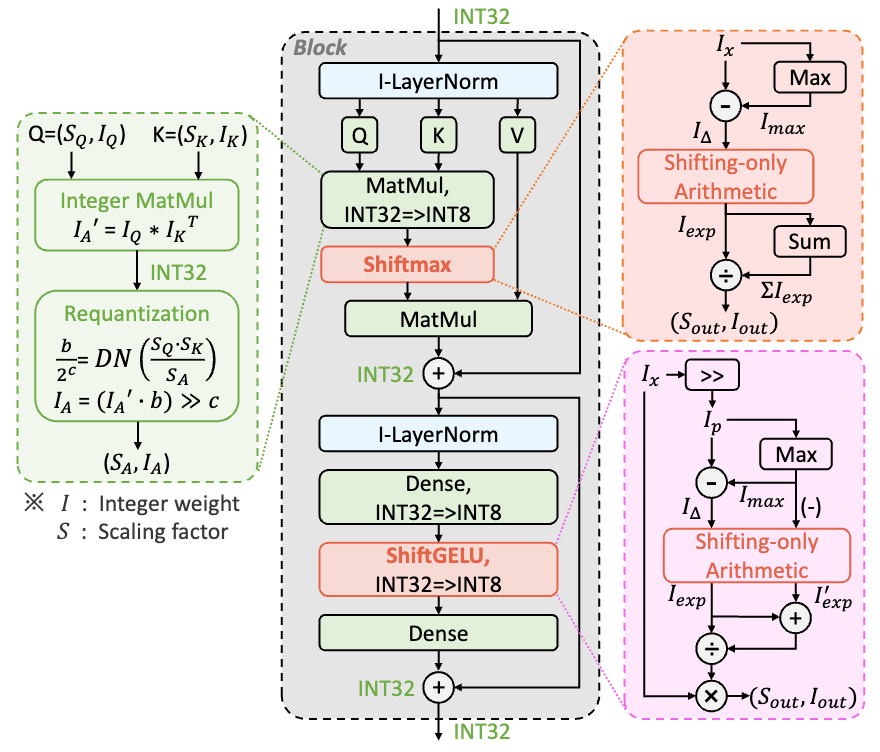}  
    \caption{I-ViT performs all calculations using integer arithmetic. Linear operations like MatMul and Dense are processed through a binary arithmetic pipeline, while custom operations like Shiftmax, ShiftGELU, and ILayerNorm handle non-linear tasks \cite{i-vit}.}
    \label{fig:i-vit}
\end{figure}

I-ViT \cite{i-vit} employs ``shiftmax," which transforms the exponential in the softmax function to base 2, leveraging the base change formula for operational efficiency through bit-shifting. Additionally, I-ViT calculates the square root in LayerNorm using a lightweight, integer-based iterative method inspired by Newton's Method, as per \cite{i-bert}, refined further with bit-shifting enhancements. For GELU approximation, ``ShiftGELU" is introduced, which uses sigmoid-based approximation and simplifies exponentiation and division to predominantly bit-shifting operations. The overview of I-ViT is shown in Fig.\textcolor{red}{\ref{fig:i-vit}}.

EdgeKernel \cite{practical} tackles the precision challenges inherent in softmax division operations by scrutinizing the selection of an ideal bit shift parameter, 
$M$, which is critical for maintaining precision while avoiding significant bit truncations. It also implements asymmetric quantization of LayerNorm inputs to the uint16 data type, balancing computational efficiency with the need to preserve the integrity of the data.

Despite avoiding data type conversions between float and integer formats, prior studies have continued to rely on high-precision multiplication and broad bit-width data storage (eg., INT32). SOLE \cite{sole} introduces E2Softmax, which processes 8-bit quantized inputs using fixed-point arithmetic. This method employs log2 quantization post-exponentiation and an approximate logarithmic division, eschewing traditional high-precision operations. Additionally, SOLE employs dynamic compression for Layer Normalization statistics via Power-of-Two Factor \cite{fq-vit}, and innovatively designs a two-stage LayerNorm Unit. This design facilitates pipelined computations and flexible batch processing, with robust error tolerance and reduced buffering and multiplication.

\subsection{\textbf{Hardware Accelerator For Quantized ViTs}} \label{sec:accelerator}

\begin{figure*}[ht]
    \centering
    \includegraphics[width=0.9\linewidth]{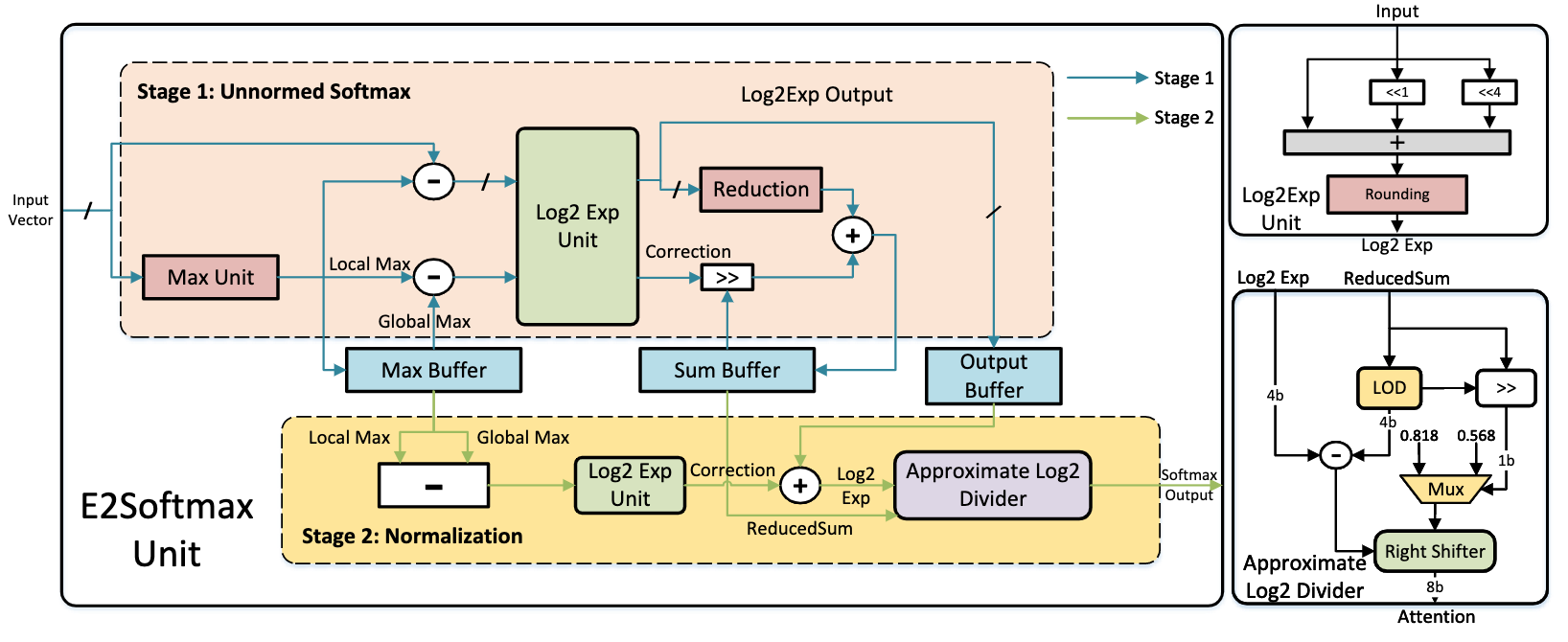}  
    \caption{E2Softmax Unit in SOLE \cite{sole}.}
    \label{fig:sole}
\end{figure*}

Designing specific hardware accelerators for quantized ViTs is essential due to their unique computational demands, such as complex attention mechanisms and unusual memory access patterns, which are not efficiently handled by general-purpose hardware. These accelerators can be optimized to manage the challenges of quantization, maintaining model performance while reducing precision, and are crucial for meeting the stringent energy efficiency and low-latency requirements of edge devices. Custom hardware solutions enable scalable and energy-efficient deployment of ViTs across various applications, ensuring real-time processing capabilities and adherence to resource constraints typical of edge computing environments. For a deeper understanding of the foundational architecture of hardware accelerators, we direct readers to \cite{survey_hardware1,survey_hardware2,survey_hardware3}. In the subsequent paragraph, we delve into the existing hardware designs on accelerating quantized ViTs with summarization in Table \ref{tab:hardware}.

VAQF \cite{vaqf} discusses optimizing Vision Transformer (ViT) models for FPGA implementation. This involves starting with a baseline using 16-bit fixed-point representations to reduce computation and storage needs, implementing data packing techniques to minimize block RAM usage and data transfer latency, and maximizing computation parallelism by determining optimal tiling sizes. The accelerator can handle both quantized and unquantized computations, with specific parameters set for unquantized layers. Lastly, parameters are fine-tuned to meet desired frame rate targets, with a compilation step determining the necessary precision for activations.

Auto-ViT-Acc \cite{auto-vit-acc} introduces an all-encompassing framework that features a mechanism for exploring design spaces, a module for modeling FPGA resource utilization, and an innovative FPGA compute engine tailored for the multi-head attention mechanism in Vision Transformers (ViTs). This framework is designed to autonomously determine the best mix of quantization bit-widths and scheme mixing ratios to achieve specified target frame rates (FPS), thereby efficiently directing the quantization process and the design of FPGA accelerators.

Huang et al. \cite{huang2023integer} implement an integer-only quantization strategy for nonlinear operations to simplify computations, enhancing compatibility with edge devices. A critical element of their approach is the versatile group vector systolic array, optimized for the efficient execution of matrix multiplication tasks. Additionally, the framework employs a unified strategy for packaging data, promoting efficient data movement and storage, and a dynamic on-/off-chip data storage management strategy, crucial for maintaining high throughput and low latency.

\begin{table}[b]
    \centering
    \resizebox{0.5\textwidth}{!}{
    \begin{tabular}{ccccc}
    \toprule
        Method & Platform & Bit-width & Model & Execution Performance \\ \midrule \midrule
        VAQF \cite{vaqf} & FPGA ZCU102 & W1A8 & Deit-Base & 861.2 GOP/S \\ 
        Auto-ViT-Acc \cite{auto-vit-acc} & FPGA ZCU102 & W8A8 + W4A8 & Deit-Base & 1181.5 GOP/S \\
        Huang el al. \cite{huang2023integer} & FPGA ZCU102 & W8A8 & ViT-Small & 762.7 GOP/s \\ 
        SOLE \cite{sole} & ASIC & W8A8 & DeiT-Tiny & - \\ \bottomrule
    \end{tabular}
    }
    \caption{Accelerator with model quantization.}
    \label{tab:hardware}
\end{table}

SOLE \cite{sole} designs custom hardware for efficient transformer inference, optimizes Softmax and LayerNorm operations. As shown in Fig.\textcolor{red}{\ref{fig:sole}}, its E2Softmax Unit avoids large lookup tables and multiplications, uses log2 quantization on exponent function outputs, and stores intermediate results in 4-bit representations, reducing memory usage. The unit also incorporates online normalization and ping-pong buffers, maximizing throughput. The AILayerNorm Unit, on the other hand, uses low-precision statistic calculation, dynamic compression, and Power-of-Two Factor quantization, employs a 16-entry lookup table for square functions and shift operations, and facilitates smooth data flow through ping-pong buffers.

\section{\textbf{Conclusions and Future Directions}}
\label{sec:conclusion}

In this paper, we systematically investigate methods for the quantization and hardware acceleration of ViTs, addressing the challenge of their deployment. ViTs are characterized by their unique self-attention module, requiring specifically tailored quantization techniques to achieve optimal compression rates without sacrificing application accuracy. Moreover, the integration of efficient computation, higher bandwidth, a limited set of operations, and opportunities for data reuse has guided the development of specialized hardware accelerators for quantized ViTs. Our survey encompasses a wide range of recent works, providing a comprehensive roadmap for the quantization of ViTs. Subsequently, we further explore the interconnections among various methods, propose directions for future research in this evolving field.

\textbf{Extremly Low-Bit:} 
The challenge of quantizing ViTs to sub-2-bit representations, as highlighted by the more than 10\% degradation in model accuracy (referenced in Table \ref{tab:acc_compare2}), underscores a critical bottleneck in the adaptation of extremely low-bit quantization for ViTs. This degradation starkly contrasts with the near-lossless performance observed in CNNs under similar quantization constraints \cite{group_net,bibench}. Continued innovation in this area could offer a path toward highly efficient ViTs deployment with minimal computational and memory footprints.

\textbf{Sub-8-bit Hardware:} As demonstrated in Table \ref{tab:hardware}, the prevailing hardware landscape is predominantly geared towards 8-bit computations. However, the minimal accuracy loss associated with 4-bit quantization algorithms, even in the context of PTQ, underscores the feasibility and potential benefits of developing accelerators tailored for sub-8-bit operations. Designing such low-bit accelerators could significantly enhance computational efficiency and throughput, reduce energy consumption, and lower hardware costs.

\textbf{Combination with Other Compression Methods:} The integration of quantization with other compression techniques such as pruning represents an underexplored yet promising avenue \cite{prune_vit}. This holistic approach to model compression can yield significant reductions in size and computational demands.

\textbf{Lower-level ViTs Quantization:} Quantizing ViTs for lower-level tasks like object detection and image generation poses distinct challenges \cite{od_quant, he2024ptqd}, underscoring the importance of specialized strategies to maintain performance. This area requires further exploration to fully harness the efficiency of quantization while ensuring the effectiveness of ViTs across a broad spectrum of applications.

\textbf{Robustness and Generalization:} Ensuring that quantized ViTs maintain robustness and generalization across diverse tasks and datasets is paramount. Future studies should investigate the impact of quantization on the model's ability to generalize and propose techniques to mitigate any adverse effects, thereby enhancing the model's utility in real-world applications. \cite{bibench}







\bibliographystyle{IEEEtran}
\bibliography{root.bib}

\end{document}